%% file: main.tex
\documentclass{article}

\usepackage[preprint]{neurips_2026}

\usepackage[utf8]{inputenc} 
\usepackage[T1]{fontenc}    
\usepackage{url}            
\usepackage{booktabs}       
\usepackage{amsfonts}       
\usepackage{nicefrac}       
\usepackage{microtype}      
\usepackage{xcolor}         
\usepackage{graphicx}
\usepackage{amsmath}
\usepackage{subcaption}
\usepackage{arydshln}
\usepackage{amsthm}
\usepackage{wrapfig}

\usepackage{hyperref}       
\usepackage[capitalize,noabbrev]{cleveref}
\newcommand{\Csq}{\texorpdfstring{C\textsuperscript{2}}{C^2}}

\theoremstyle{plain}
\newtheorem{theorem}{Theorem}[section]
\newtheorem{proposition}[theorem]{Proposition}

\theoremstyle{definition}

\theoremstyle{remark}

\usepackage{amssymb}
\usepackage{mathtools} 

\title{Compositionality Emerges in a Narrow Depth–Connectivity Regime: Architecture Constraints and Solution Manifolds}

\author{
Dat H. Do$^{1}$\thanks{Corresponding to 22dat.dh@vinuni.edu.vn},
Rushi Shah$^{1}$,
Duc V. Le$^{2}$,
and Dianbo Liu$^{1}$ \\
National University of Singapore$^{1}$ \\
University of Twente$^{2}$
}

\begin{document}

\maketitle
\begin{abstract}
Compositionality is believed to be the foundation for generalization, enabling models to reuse meaningful primitives in novel combinations. Yet, models trained with standard gradient-based optimization rarely, and often only weakly, exhibit compositional internal structure, and it remains unclear how or why such compositionality forms. In this work, we show that compositionality emerges in a narrow connectivity-depth sweet spot. Along the connectivity axis, compositionality only appears in some specifically sparse networks, heavily depends on which connections remain rather than on weights' sparsity alone. Along the depth axis, compositionality emerges within a narrow, target-dependent regime, peaking at specific depths, while both shallower and deeper networks fail. When either the depth or connectivity condition is violated, gradient descent silently converges to fractured solutions rather than compositional ones. To discover and exploit this emergence, we introduce (i) similarity-based pruning (SP) to recover compositional connectivity and (ii) a heuristic depth predictor to estimate where compositionality is most likely to appear. Finally, we support these empirical findings with a theoretical framework based on compositional sparsity, volume-ratio arguments, and feature-interference bounds, explaining why compositional solutions are reachable only in a narrow depth–connectivity regime.
\end{abstract}

\input{sections/1_intro.tex}

\input{sections/4_empirical_result.tex}

\input{sections/3_theory.tex}
\input{sections/6_implications.tex}
\input{sections/6_related_works.tex}
\input{sections/7_conclusion.tex}

\bibliography{ref}
\bibliographystyle{unsrt}

\input{sections/5_bis_appendix.tex}

\input{sections/appendix.tex}


\end{document}

%% file: sections/1_intro.tex
\section{Introduction}

Modern foundation models such as large language models (LLMs), vision–language models (VLMs), and diffusion models exhibit strong capabilities in reasoning behavior and producing photorealistic images and videos, yet generalization remains fragile under distribution shift, with failures concentrated in compositional understanding. Small, task-irrelevant changes can trigger sharp behavioral differences. For LLMs, seemingly meaning-preserving prompt edits, minor changes in formatting or phrasing, can noticeably alter behavior \cite{lampinen2025generalization, berglund2023reversal}.  For VLMs, small visual prompt perturbations (e.g., marker color \cite{Feng2025VisuallyPB}) can dramatically change outcomes and even reshuffle model rankings across splits \cite{xu2025vp}. Diffusion models similarly synthesize photorealistic scenes but still struggle with attribute binding, object relations, and counting, motivating compositional benchmarks like T2I-CompBench \cite{huang2023t2i, huang2025t2i}, and object-centric evaluators like GenEval \cite{ghosh2023geneval,kamath2025geneval}. These observations suggest that strong model capabilities do not guarantee robust generalization, and compositionality is a necessary foundation for generalizing reliably beyond observed distributions.


A mechanistic view attributes such brittleness to representational factors. In particular, \cite{elhage2022toy} shows that neurons become polysemantic, responding to multiple unrelated concepts within a single unit. Such entangled activations stand in direct contrast to compositional representations built from reusable and semantically stable primitives. From this perspective, compositionality and monosemanticity are tightly linked: \textbf{monosemantic features} - defined as internal representations that correspond to semantically meaningful factors across contexts, while \textbf{compositional circuits} - defined as subnetworks that combine these representations to produce the output, organize them into reliable computations.

\begin{figure*}[t]
    \centering
    \includegraphics[width=0.98\linewidth]{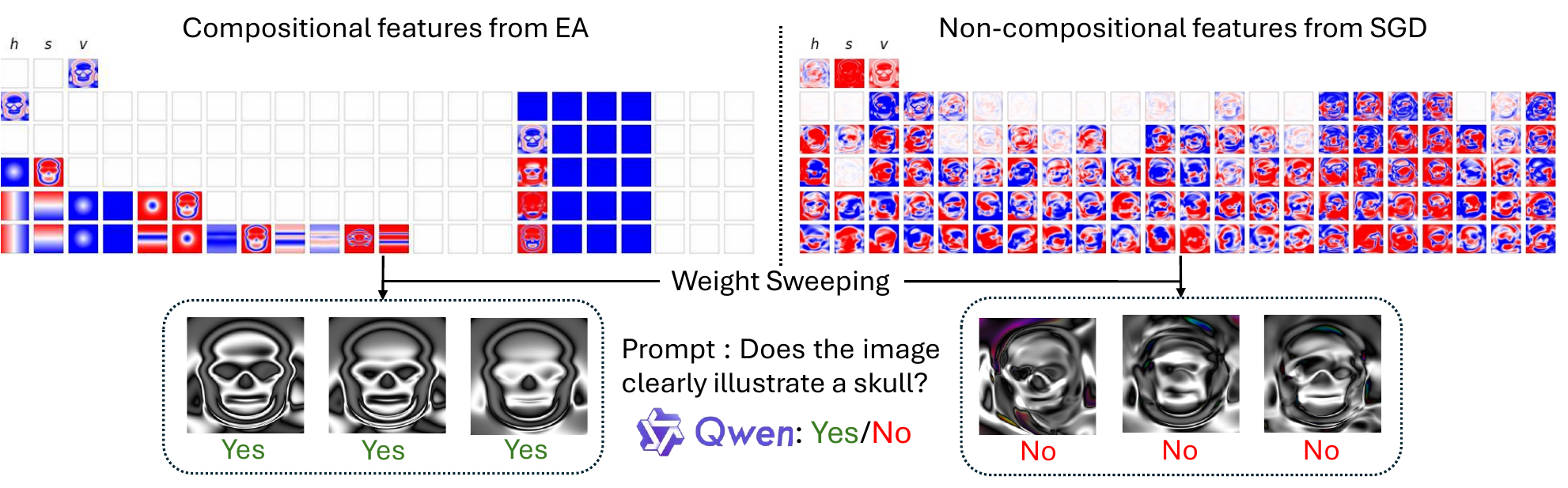}
    \caption{\textbf{Comparing internal structure (Red=-1, White=0, Blue=1)}: an evolutionary algorithm (EA) \cite{secretan2008picbreeder} setup can yield factorized, reusable intermediate features, whereas an SGD-trained network \cite{kumar2025fer} often exhibits fragmented, entangled ones. We quantify compositionality via \emph{weight sweeping}: perturb each nonzero parameter by noise~$\delta$ and query a calibrated VLM judge under a fixed prompt to assess whether the image still depicts the target concept (``skull''). The proportion of affirmative judgments defines the compositional score. }
    \vspace{-1em}
    \label{fig:problem_statement}
\end{figure*}

Even when a task admits clean compositional circuits, gradient-based training fails to find them. The Fractured Entangled Representations (FER) hypothesis \cite{kumar2025fer} sharpens this: gradient-based training tends to converge to \textbf{fractured entangled representations} - defined as features fail to align with any stable, interpretable meaning and concepts are distributed across many overlapping neurons. Meanwhile, alternative procedures like evolutionary algorithms can uncover Unified Factored Representations (UFR) that isolate reusable components, shown in Figure~\ref{fig:problem_statement}. This gap between (i) the existence of compact, modular implementations and (ii) what vanilla backpropagation reaches in practice implies that training dynamics and architectures may systematically favor FER that are more brittle under shift, motivating our central question:
\textbf{\emph{Can monosemanticity and compositionality emerge during gradient-based backpropagation training, and if so, when do they emerge?}}

We approach this question as a structural reachability problem governed jointly by \textbf{depth} - the number of layers in the network, and \textbf{connectivity} - the way neurons are wired together, sparsely or fully connected. To make the phenomena measurable and reproducible, we introduce EM\Csq-Bench, a perturbation-based evaluation suite exposing monosemantic features. In particular, we study connectivity by introducing a new pruning algorithm that uncovers compositional subnetworks. We then use that algorithm to study how depth affects compositionality.

Our findings reveal a pronounced peak in compositionality at intermediate depths, and that naive capacity control or uniform pruning is insufficient: \emph{which} connections remain matters more than \emph{how many}. Additionally, prior evidence for UFR is limited, since the \textbf{Picbreeder's artifacts/images}-skull, butterfly, apple images in Appendix~\ref{ap:original_picbreeder}- are synthetic outcomes of a human-curated process. We therefore show that our method transfers to images where the underlying compositional solutions are not known in advance. Finally, we provide a theoretical framework based on compositional sparsity, volume-ratio arguments, and feature-interference bounds to explain why compositional solutions are reachable only within a narrow depth--connectivity regime. Our contributions are fourfold:
\begin{enumerate}
    \item We formulate emergent compositionality as a structural reachability problem by empirically mapping its dependence on the intertwinement of depth and connectivity.
    \item We release a set of standardized tasks and an evaluation suite, EMC\(^2\)-Bench, that enables consistent quantitative comparisons of compositionality across different networks.
    \item We introduce two minimal interventions: a target-complexity depth predictor that anticipates depth regions where compositionality peaks, and Similarity-based Pruning (SP), a structure-aware pruning rule that induces connectivity patterns conducive to compositional circuits.
    \item We provide a theoretical framework based on compositional sparsity, volume-ratio arguments, and feature-interference bounds, explaining why compositional solutions are reachable only within a narrow depth--connectivity regime.
\end{enumerate}

%% file: sections/4_empirical_result.tex
\section{Gradient Descent finds Compositionality in a narrow Depth and Connectivity regime}

\subsection{Preliminaries}
\label{sec:preliminaries}

Picbreeder \citep{secretan2008picbreeder} is an open-ended system in which human users iteratively breed images by selecting preferred candidates, while the evolutionary algorithm NEAT \cite{stanley2002efficient} mutates and recombines the underlying networks. \cite{kumar2025fer} leverages this setting as a rare point of comparison for representation learning: Picbreeder routinely discovers compositional pattern-producing networks (CPPN) whose internals appear \emph{unified and factored}, providing an explicit example of a compositional solution that produces a coherent object. The original NEAT CPPNs (details in Appendix~\ref {ap:imp_details}) can be ported to computationally equivalent dense MLPs via layerization, yielding an architecture compatible with standard backpropagation while preserving the original computation \cite{kumar2025fer}. Crucially, \cite{kumar2025fer} also shows that SGD on this supervised objective can closely match the target image in output space. Yet, the learned solution exhibits FER, where underlying regularities and symmetries are not cleanly reused throughout the network. Small weight perturbations tend to break semantics incoherently, as shown in Figure~\ref{fig:problem_statement}. We restrict our study to MLP architectures to maintain a fully inspectable setting. We do not experiment on other architectures such as CNNs or Transformers, as the existence and characterization of compositional solutions in those architectures remain unclear, making them unsuitable for controlled analysis of compositional emergence.

\subsection{Emergent Compositionality}
\label{sec:picbreeder_adam}
\begin{figure}[!ht]
    \centering

    \begin{minipage}[t]{0.48\linewidth}
        \centering
        \includegraphics[width=\linewidth]{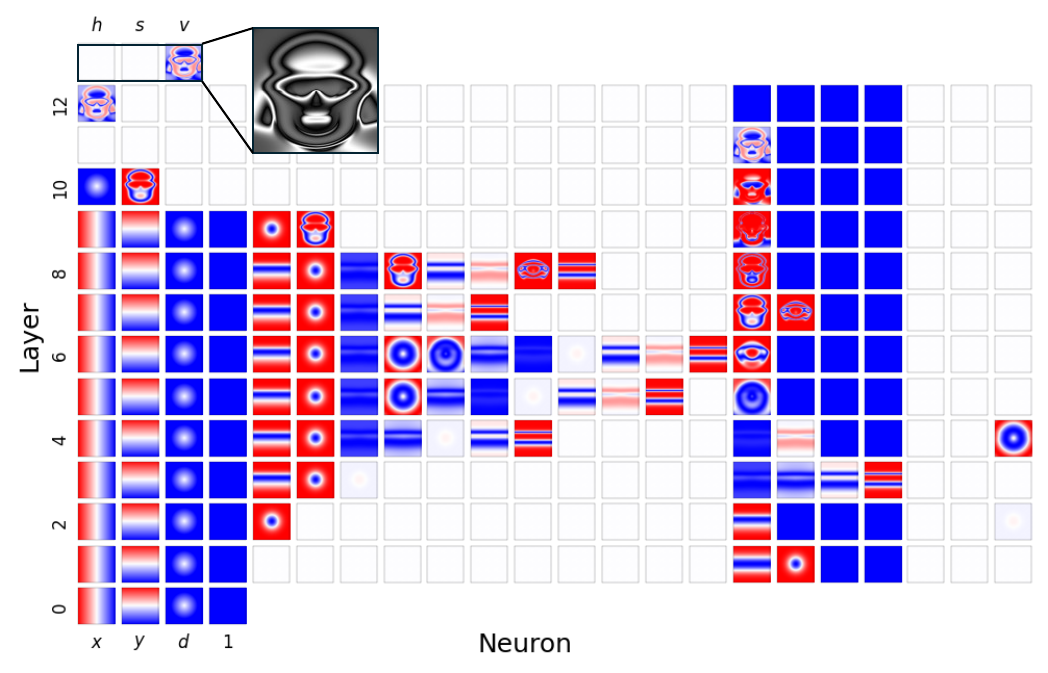}
    \end{minipage}
    \hfill
    \begin{minipage}[t]{0.48\linewidth}
        \centering
        \includegraphics[width=\linewidth]{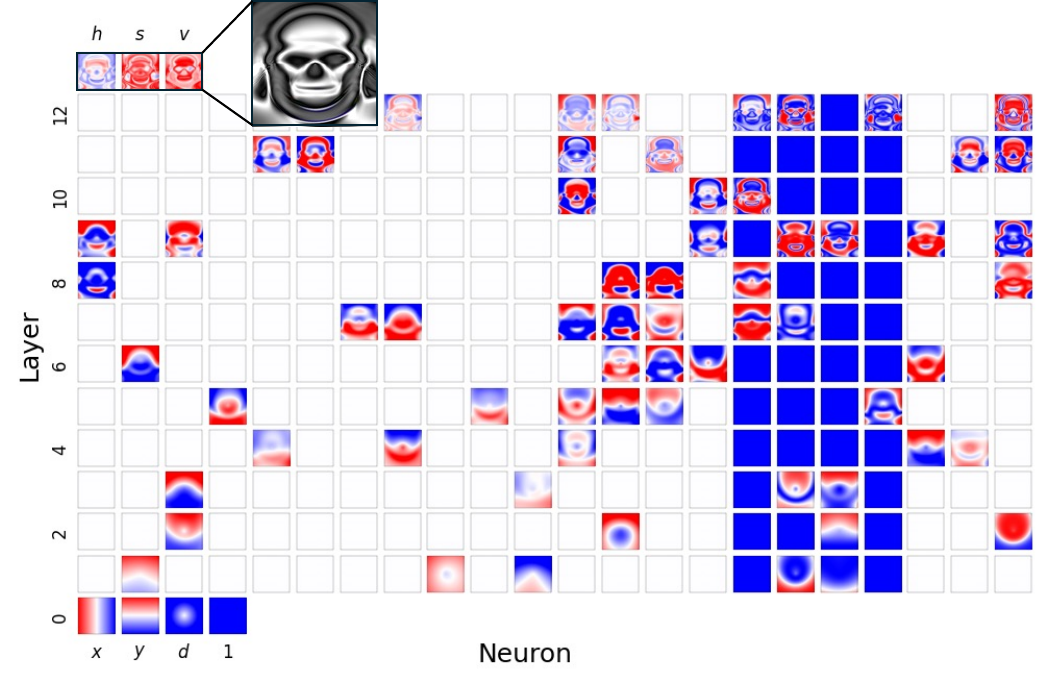}
    \end{minipage}

    \caption{\textbf{Architectural bias shapes compositionality (Red=-1, White=0, Blue=1).} The MLPs takes pixel coordinate inputs $(x,y,d=\sqrt{x^2+y^2},1)$ and predicts $(h,s,v)$, which are converted into the RGB image. Each square box visualizes the output/activation map of a single neuron over the image grid. \textbf{Left}: Preserving the NEAT sparse wiring and retraining the MLP yields partially compositional intermediate features. \textbf{Right}: Similarity-based Pruning (Section~\ref{sec:pruning}) produces a subnetwork that exhibits emergent compositionality,  pruning from 5477 to 343 weights. }
    \label{fig:arch_pruning}
\end{figure}

The Picbreeder's models exhibit strong compositionality and monosemanticity. Noticeably, they all appear unusually sparse and structurally specialized, raising the possibility that their distinctive architecture is what enables these representations. To isolate the role of this structure, we take the Picbreeder's network and re-initialize it by preserving its 0/1-valued weights (which encode the specialized wiring pattern) while re-sampling the remaining trainable parameters before applying gradient-based optimization. Figure~\ref{fig:arch_pruning} (left) visualizes the results of this experiment, suggesting that emergence is possible but highly contingent. With the right architectural bias and hyperparameters, Adam can yield partially compositional internal organization, including a striking tendency toward vertically symmetric intermediate features. Yet this structure is fragile. Despite the existence of a clearly compositional global minimum (the original Picbreeder's models), optimizers never reach it as MLPs involve non-convex loss landscapes, trapping algorithms in local minima. Hence, strong compositionality is not the default outcome of SGD, but an emergent property that depends on architectural bias, strengthened by an optimizer, which is capable of finding the global minima.
\subsection{Connectivity as a Regime Control}
\label{sec:connectivity}


Given that the emergence of monosemantic features and compositional structure is highly contingent on architectural bias, as shown in Section~\ref{sec:picbreeder_adam}, we next ask whether we can recover an effective architecture that makes such structure reliably accessible. One possibility is that SGD already learns partially compositional features, but that they are obscured by redundancy. To test this, we apply similarity-based pruning (SP), which prunes neurons that are highly similar to others within the same layer (see Section~\ref{sec:pruning} for details of SP). Figure~\ref{fig:arch_pruning} (right) shows that SP can collapse a dense, entangled solution into a much smaller circuit. The resulting connectivity is markedly sparser, and the remaining feature maps display clearer spatial motifs and notably consistent vertical symmetries, suggesting the computation is beginning to organize into more stable, reusable primitives, an observable signature of emergent compositionality. Importantly, when we prune to the same parameter count using other methods (see Appendix~\ref{ap:prune}), we do not observe comparable compositionality. This suggests that how connections are organized matters more than the sheer number of connections.

\subsection{Depth Sets the Regime Boundary}
\label{sec:depth_effect}
\begin{wrapfigure}{r}{0.51\linewidth}
    \centering
    \vspace{-1em}
    \includegraphics[width=\linewidth]{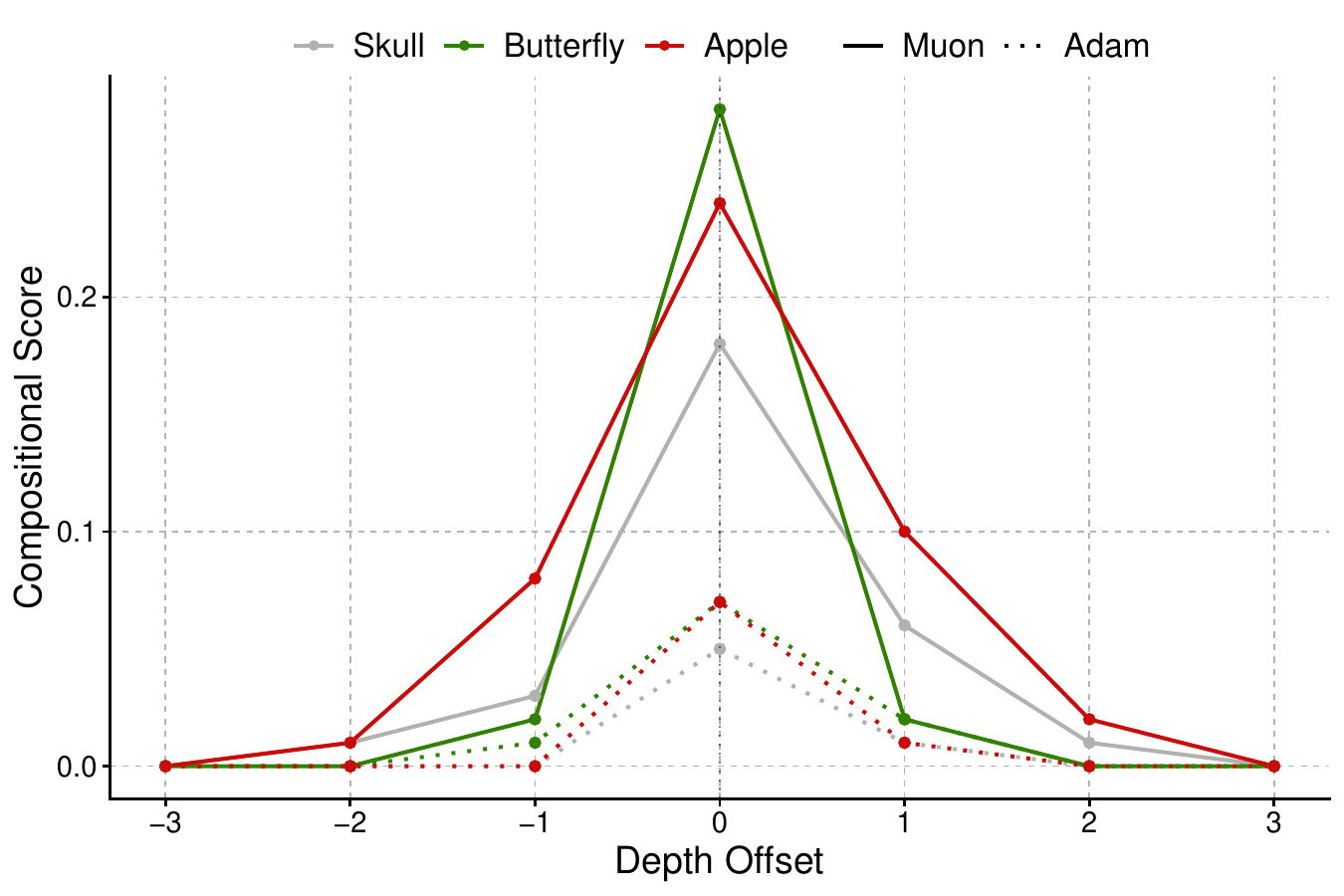}
    \caption{\textbf{Compositional score vs depth offset on Picbreeder artifacts.} We vary only the network depth around a reference depth while keeping pruning and training settings fixed. Each data exhibits a peak at a target-specific depth, while shallower and deeper networks show reduced modularity.}
    \label{fig:depth}
    \vspace{-1em}
\end{wrapfigure}

The original Picbreeder CPPNs (see Appendix~\ref{ap:original_picbreeder}) already exhibit different depths across artifacts, suggesting that compositional structure may require a target-specific architectural depth rather than a single universal choice. Hence, we ask whether compositionality is sensitive to depth even when all other factors are held fixed. Other than depth, all implementation settings, including applying the same SP strategy, are kept identical across conditions. We adjust the number of layers around a reference depth and denote this change as a depth offset \(\Delta d \in \{-3,\ldots,3\}\), and compute the compositional score. Figure~\ref{fig:depth} reports the resulting compositional score as a function of \(\Delta d\) for three representative objects (qualitative results are in the Appendix.~\ref{ap:depth}). Here, the compositional score measures the extent to which learned weights remain semantically invariant under perturbations; implementation details of this measurement are given in Section~\ref{sec:vllm_eval}. Across all targets and optimizers, compositionality exhibits a clear depth dependence: the score peaks at a specific depth and drops rapidly when the network is made either shallower or deeper. Notably, the optimal depth is target-dependent, with different artifacts exhibiting their peak compositional score at different depths. This suggests the existence of a depth at which the model most naturally organizes computation into reusable parts. Overall, depth is not merely a capacity knob: it strongly shapes whether the learned solution is implemented as a compact set of semantically stable components.

\subsection{Depth \texorpdfstring{$\times$}{x} Connectivity: The Compositional Regime}

Taken together, our results suggest that compositionality is not determined by depth or connectivity in isolation, but by their interaction. Connectivity controls whether features are forced to specialize, while depth controls whether the network has the right number to factor computation into reusable sub-functions. Compositionality, therefore, emerges only within a narrow depth-connectivity regime.

\section{Methodologies}

\subsection{EM\texorpdfstring{\Csq}{Csq}-Bench: Benchmarking \underline{E}megent \underline{M}onosemanticity \& \underline{C}ompositional \underline{C}ircuits}
\label{sec:vllm_eval}
To better understand when monosemanticity and compositionality emerge, we first need a reliable evaluator. We quantify compositionality by asking a simple question: \emph{Does a small, localized change to a single parameter preserve the global semantic identity of the generated image?} As illustrated in Figure~\ref {fig:problem_statement}, perturbing some weights yields an output that is still recognized as the same object/category (\texttt{Yes}), while perturbing other weights causes a semantic drift (\texttt{No}). Concretely, we iterate over all non-zero parameters $w \in \mathcal{W}_{\neq 0}$ and create a perturbed copy $w' = w + \delta$ while keeping all other parameters fixed.  We then run the network to obtain the perturbed output image and query a VLM. If the VLM returns \texttt{Yes}, we mark $w$ as \emph{compositional}. Finally, we define the compositional score as the fraction of non-zero parameters passing the semantic-invariance test:
\begin{equation}
\mathrm{CompositionalScore}
= \frac{N_{\mathrm{comp}}}{N_{\neq 0}}
\label{eq:compositional_score}
\end{equation}
with $N_{\mathrm{comp}}$ is the number of compositional weights and $N_{\neq 0}$ is the total number of non-zero weights. While this metric relies on a VLM evaluator and is not a perfect measure of semantic invariance, we observe that the quantitative results in Table~\ref{tab:compositional_scores} align with qualitative analyses in the Appendix~\ref{ap:other_pruning} under relative comparison, suggesting that EMC$^2$-Bench is sufficiently reliable for our purposes.

\subsection{SP: Similarity-based Pruning}
\label{sec:pruning}
Our central premise is that compositional structure emerges when intermediate representations allocate distinct and reusable neurons to different semantic factors, rather than distributing the same factor across multiple redundant neurons. We therefore prune directly at the neuron level. Specifically, we follow the iterative prune--retrain paradigm: each round prunes redundant neurons, then fine-tunes the remaining weights to recover performance before repeating the process. In detail, we apply this procedure independently to each layer: for a given layer $i$, we extract the layer's neuron-wise feature maps $\{\mathbf{f}^{(i)}_k\}$ and compute the pairwise cosine similarity matrix $S_{kj}$ over these features. Then, for each neuron $k$, we select its most similar neuron $p(k)$  and retain only redundancy pairs whose cosine similarity exceeds a threshold $\tau$:
\begin{equation}
\begin{aligned}
S_{kj} &=
\frac{\langle \mathbf{f}^{(i)}_k,\mathbf{f}^{(i)}_j\rangle}
{\|\mathbf{f}^{(i)}_k\|_2\,\|\mathbf{f}^{(i)}_j\|_2},
\qquad
p(k) &= \arg\max_{j\neq k} S_{kj},
\qquad
\mathcal{P}_i &= \{\, S_{k,p(k)} > \tau \,\}.
\end{aligned}
\label{eq:pairs-horizontal}
\end{equation}
To determine a pruned set $\mathcal{R}_i$ for layer $i$, we then iterate over redundancy pairs $(k,j)\in\mathcal{P}_i$: if either neuron has already been pruned ( $k\in\mathcal{R}_i$ or $j\in\mathcal{R}_i$), skip the pair; otherwise, randomly select one of the two neurons uniformly and add it to $\mathcal{R}_i$. Finally, for every pruned neuron $r\in\mathcal{R}_i$, we structurally disconnect it by zeroing both incoming and outgoing weights: $(W_{i-1})_{[:,r]}\leftarrow \mathbf{0}, (W_i)_{[r,:]}\leftarrow \mathbf{0}.$ After pruning, we continue to fine-tune the remaining sub-network without any weight re-initialization. 
\subsection{Compositional Depth Predictor}
\label{sec:heuristic_depth_predict}
\begin{wrapfigure}{r}{0.51\linewidth}
    \centering
    \vspace{-1em}
    \includegraphics[width=\linewidth]{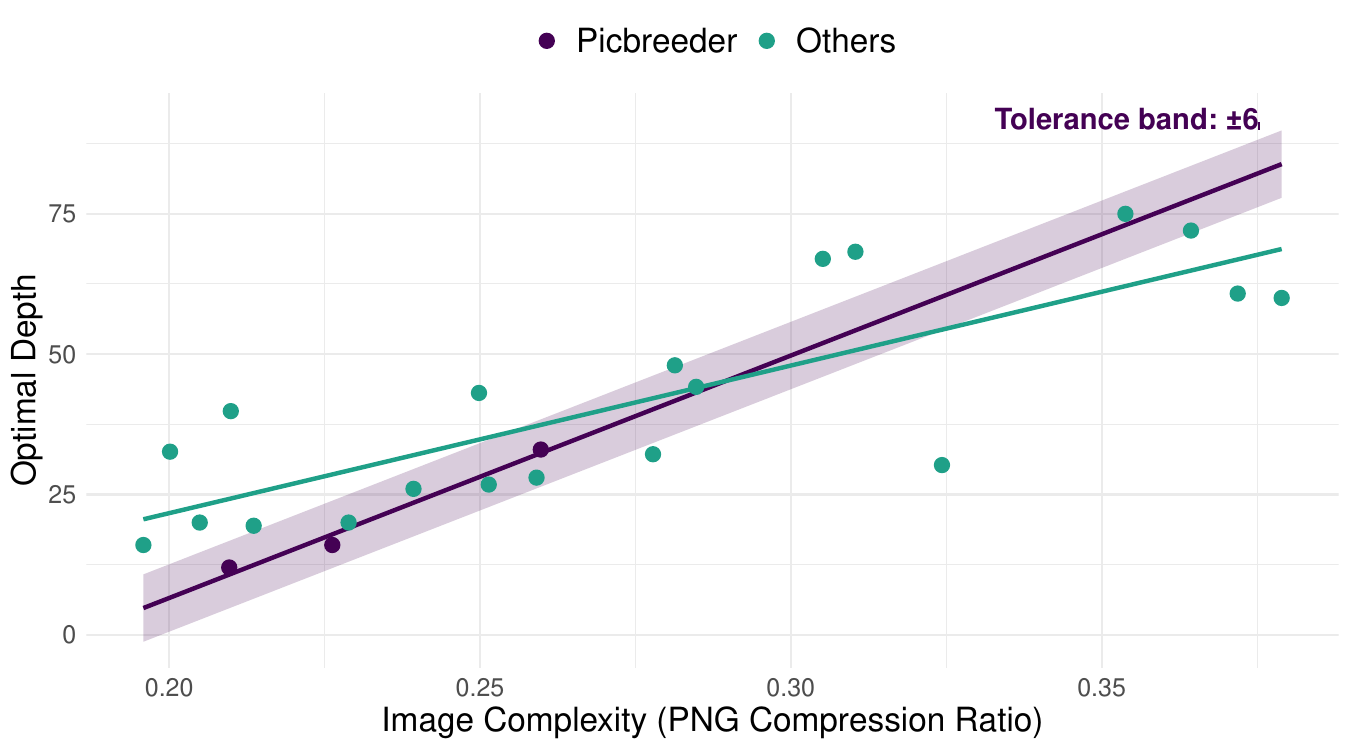}
    \caption{\textbf{Image complexity predicts optimally compositional depth.} We compute the PNG compression ratio for each target image and compare it with the empirically optimal depth found by depth sweep. Higher image complexity tends to correlate with larger optimal compositional depth.}
    \label{fig:depth_complexity}
    \vspace{-2em}
\end{wrapfigure}
Section.~\ref{sec:depth_effect} shows that compositionality peaks at a target-specific optimal compositional depth, implying there is no single depth that works best across images. Consequently, when optimizing for targets beyond Picbreeder's artifacts, we want a way to efficiently locate a good depth without performing an exhaustive sweep. Hence, we introduce the use of image complexity, measured by the PNG compression ratio of the ground-truth image, as a lightweight heuristic: visually richer images that compress less efficiently should have more compositional subfunctions. 
The PNG compression ratio is defined as $\mathrm{CR}(x) = \frac{\mathrm{Size}_{\text{PNG}}(x)}{\mathrm{Size}_{\text{raw}}(x)}$, where $\mathrm{Size}_{\text{raw}}(x) = H \times W \times C$.

Figure~\ref{fig:depth_complexity} plots Picbreeder's artifacts (in purple dots) and reveals an approximately linear relationship between the optimal depth and PNG compression ratio; we fit a linear trend on Picbreeder (purple). For new images (``Others''), we extrapolate a predicted depth from this fitted line and then continue adjusting the depth until we recover the empirical optimum. We additionally fit a separate linear regressor directly on the non-Picbreeder images (green) and observe that this fitted line deviates modestly from the Picbreeder line. The opaque band of $\pm 6$ layers around the predictor is used to evaluate the heuristic, indicating that the predictor typically lands near the optimum and can narrow the depth search space. However, even under the new fitted line, the measured optimal depths remain noisy, with substantial variance, suggesting additional factors are needed for better precision.

\section{Experiments}
\label{sec:experiments}
\subsection{Quantitative Results}

\begin{wraptable}{r}{0.51\linewidth}
\vspace{-1em}
\centering
\scriptsize
\setlength{\tabcolsep}{3pt}
\resizebox{\linewidth}{!}{
\begin{tabular}{lccc}
\toprule
\textbf{Method} & \textbf{Skull} & \textbf{Butterfly} & \textbf{Apple} \\
\midrule
Lottery Ticket Hypothesis & 0.00 & 0.00 & 0.00 \\
Wanda & 0.00 & 0.00 & 0.00 \\
LLM-Pruner & 0.00 & 0.00 & 0.00 \\
\hdashline
SP (Adam) & 0.05 & 0.07 & 0.07 \\
SP (Muon) & 0.12 & 0.21 & 0.21 \\
SP (Muon, NS step=20) & 0.18 & 0.28 & 0.24 \\
\hdashline
Picbreeder & 0.63 & 0.86 & 0.57 \\
\bottomrule
\end{tabular}
}
\caption{\textbf{Quantitative results on EM\Csq-Bench across different pruning methods.} SP is the only method with non-zero scores, improving with better optimizers. Corresponding qualitative results are in Appendix~\ref{ap:full_viz} and \ref{ap:other_pruning}.}
\label{tab:compositional_scores}
\vspace{-1em}
\end{wraptable}

We quantitatively evaluate pruning methods on EMC$^2$-Bench to measure whether they recover semantic-invariant compositional parameters. Table~\ref{tab:compositional_scores} shows that SP outperforms standard pruning methods under a fixed parameter budget. Baselines (Lottery Ticket Hypothesis \cite{frankle2018lottery}, Wanda \cite{sun2023simple}, and LLM-Pruner \cite{ma2023llm}) all score 0, indicating no recovery of semantic-invariant parameters under weight sweeping (see Appendix~\ref{ap:prune} for qualitative results). In contrast, SP consistently achieves non-zero scores, which improve with better optimization. Despite identical parameter counts, the differing internal organization strengthens that connectivity, not just size, drives monosemanticity and compositionality (Section~\ref{sec:connectivity}). Optimization also plays a key role. Results in Appendix~\ref{ap:final_train_loss} and ~\ref{ap:lr} show that Muon achieves lower training loss than Adam, especially with more Newton--Schulz steps. Since architecture and capacity are unchanged, this reflects improved optimization rather than increased model size. The final training loss gap aligns with the quantitative score gap, indicating that better minima lead to stronger compositionality.

\subsection{Out-of-Domain Targets}
\begin{figure*}[!ht]
    \centering
    \includegraphics[width=\linewidth]{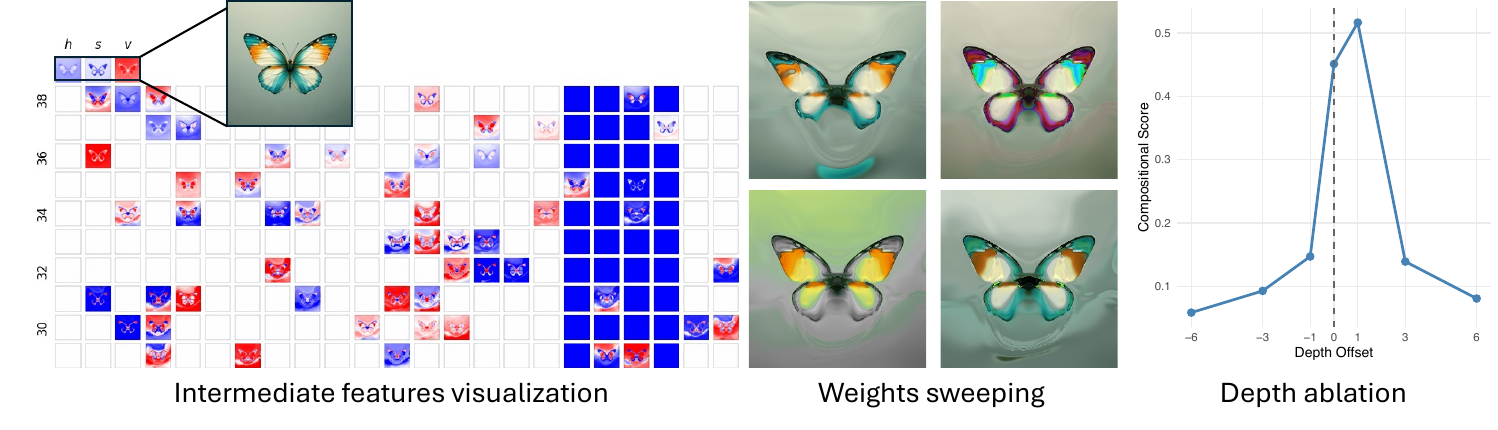}
    \caption{\textbf{Qualitative result of out-of-domain targets beyond Picbreeder.} We apply SP and heuristic depth search to images whose underlying compositional structure is unknown. The model exhibits monosemantic intermediate features, meaningful output changes under weight sweeping, and a depth-ablation peak near the predicted depth. More results are included in the Appendix~\ref{ap:synthetic}.}
    \label{fig:synthetic}
\end{figure*}
Combining SP with heuristic depth search lets us study monosemantic features and compositional circuits across images. Figure~\ref{fig:synthetic} shows that these properties emerge beyond Picbreeder artifacts. Across all three panels, the models show consistent compositional signatures: monosemantic intermediate feature maps with vertical symmetry, meaningful output variation under weight sweeping, and a depth-ablation peak near the predicted $\hat{d}$. These results suggest that networks can discover decomposable structure without prior knowledge of the target decomposition. Additional object-level and natural-scene probes in the Appendix~\ref{ap:synthetic} show partial object-level compositionality on a multi-object scene (Figure~\ref{fig:spatial_compositionality}) and weaker, but still non-vanishing, reconstruction and compositionality under substantial background variation (Figures~\ref{fig:background} and~\ref{fig:nobackground}).

%% file: sections/3_theory.tex
\section{Theoretical Framework}
\label{sec:theory}

The compositional sweet spot documented in Section~\ref{sec:experiments} raises
two distinct questions. First, does there \emph{exist} a network of width $W$ and depth $L$ that represents the target compositionally? Second, does gradient-based training \emph{find} such a network with non-trivial probability? Classical approximation theory addresses only the first question. This section recalls the relevant background in Section~\ref{sec:theory:background}, then introduces our two contributions: a combinatorial volume-ratio argument that explains \emph{why} the sweet spot is sharply localised in $(W,L)$ space (Section~\ref{sec:theory:volume}), and a mechanistic account of \emph{how} stochastic gradient methods are biased towards the compositional basin at that location (Section~\ref{sec:theory:sgd}). Section~\ref{sec:theory:implications} revisits the empirical phenomena of Section~3 through this
lens.

\subsection{Background: Compositional Sparsity}
\label{sec:theory:background}

We adopt the compositional-sparsity framework of
\cite{Danhofer2025PositionAT}. A target $f:\mathbb{R}^d\to\mathbb{R}$ is
\emph{compositionally sparse} if it factors through a directed acyclic graph
$G^\star=(V,E)$ with $|V|=\operatorname{poly}(d)$ and bounded local arity
$d_{\max}(G^\star)=\max_{v\in V} d_v = O(1)$, where $d_v=\deg_{\mathrm{in}}(v)$.
Writing $s_v$ for the smoothness at node $v$ and
$s_{\min}=\min_{v\in V} s_v$, the deep approximation rate satisfies
\begin{equation}
\label{eq:deep-rate}
N_{\mathrm{deep}}(\varepsilon;G^\star)
\;=\;
O\!\left(\sum_{v\in V}\varepsilon^{-d_v/s_v}\right)
\;\leq\;
\operatorname{poly}(d)\cdot
O\!\left(\varepsilon^{-d_0/s_{\min}}\right),
\end{equation}
whereas a shallow approximator of a generic $f\in W_s^d$ obeys
$N_{\mathrm{shallow}}(\varepsilon)=O(\varepsilon^{-d/s})$, exponential in the ambient dimension. The relevant architectural quantity is therefore the induced dependency graph $G(W)=(V(W),E(W))$ of a parameterised network, with $(i\to j)\in E(W)\iff W_{ji}\neq 0$; deep approximation is efficient precisely when $d_{\max}(G(W))$ matches $d_{\max}(G^\star)$.

\paragraph{Existence is not emergence.} Equation~\eqref{eq:deep-rate} establishes that compositional minimisers exist and are statistically parsimonious, but it is silent on whether SGD reaches them. The remainder of this section closes that gap.

\subsection{The Volume-Ratio Argument}
\label{sec:theory:volume}

Let $\mathcal{M}_\varepsilon(W,L)\subset\mathbb{R}^p$ denote the set of parameter configurations of an MLP with width $W$ and depth $L$ that achieve training loss $\leq\varepsilon$, and let $\mathcal{M}_\varepsilon^{\mathrm{comp}} (W,L)\subseteq\mathcal{M}_\varepsilon(W,L)$
denote the subset whose induced dependency graph satisfies
$d_{\max}(G(\theta))\leq d_0$. We define the \emph{compositional volume ratio}
\begin{equation}
\label{eq:volume-ratio}
\rho(W,L)
\;=\;
\frac{\mathrm{vol}\!\left(\mathcal{M}_\varepsilon^{\mathrm{comp}}(W,L)\right)}
     {\mathrm{vol}\!\left(\mathcal{M}_\varepsilon(W,L)\right)},
\end{equation}
which measures the prior probability, under uniform sampling on
$\mathcal{M}_\varepsilon$, of landing in a compositional basin. The next two propositions characterise $\rho$ as a function of architecture; full proofs are deferred to Appendix~\ref{app:proofs}.

\begin{proposition}[Width dilution]
\label{prop:width}
Fix a target requiring $P$ primitives of arity $d_0$, and a depth
$L\geq L^\star$ (defined in Proposition~\ref{prop:depth}). Then
\begin{equation}
\label{eq:width-bound}
\rho(W,L)
\;\leq\;
\frac{\binom{W}{P}\,P!}{W^{P(L-1)}}
\;=\;
\frac{W!}{(W-P)!\,W^{P(L-1)}}.
\end{equation}
In particular, for any fixed $L\geq 2$,
$\rho(W,L)=O\!\left(W^{-P(L-1)}\right)\to 0$ as $W\to\infty$.
\end{proposition}


\begin{proposition}[Depth non-monotonicity]
\label{prop:depth}
Let $L^\star=\lceil \log_2(P+1)\rceil$ denote the minimum depth required to realise a balanced binary composition tree over $P$ primitives. Then
\begin{enumerate}
\item For $L<L^\star$, $\mathcal{M}_\varepsilon^{\mathrm{comp}}(W,L)=\emptyset$
and $\rho(W,L)=0$.
\item Subject to $W\geq W_{\min}=Pd_0$, $\rho$ is maximised at $L=L^\star$.
\item For $L>L^\star$,
$\rho(W,L)\leq \rho(W,L^\star)\cdot W^{-P(L-L^\star)}$,
which decreases exponentially in the excess depth $L-L^\star$.
\end{enumerate}
\end{proposition}

Together, Propositions~\ref{prop:width}--\ref{prop:depth} make a
non-vacuous prediction: $\rho$ is concentrated on the corner
$(W_{\min},L^\star)$ of the architecture grid and decays in every direction. The denominator $W^{PL}$ in~\eqref{eq:width-bound} is a coarse upper bound; quotienting by the layerwise permutation symmetry $S_W^L$ would tighten it, but the qualitative prediction $\rho\to 0$ as $W\to\infty$ and the location of the maximum are preserved.

\subsection{Why SGD Finds the Sweet Spot}
\label{sec:theory:sgd}

Proposition~\ref{prop:width} characterises the prior measure of compositional basins but does not by itself establish convergence: a basin can have vanishing volume yet still attract a large fraction of trajectories if it is sufficiently flat or if its gradient field is sufficiently aligned. We identify three complementary mechanisms, in decreasing order of formal strength, by which SGD is biased towards the compositional basin at the sweet spot.

\paragraph{Mechanism 1: forced uniqueness.} At $(W_{\min},L^\star)$,
Proposition~\ref{prop:width} predicts $\rho\!\to\!1$ because the bound is saturated: there is essentially one non-permutation-equivalent basin compatible with $\varepsilon$-loss. Any trajectory that interpolates the data must therefore terminate in that basin, regardless of initialisation or optimiser.

\paragraph{Mechanism 2: basin flatness.} Even when wider networks admit compositional basins, those basins are \emph{flatter} than their fractured-entangled counterparts. Compositional solutions decompose into $P$ independent primitive sub-circuits, so a perturbation of one primitive's weights leaves the others unchanged and loss-flat directions abound. Fractured solutions, by contrast, distribute each primitive's computation across many cooperating neurons; a small perturbation of any one neuron breaks the co-activation pattern and incurs loss. Formally,
\begin{equation}
\operatorname{tr}\!\left(\nabla^{2}\mathcal{L}\right)\Big|_{\theta_{\mathrm{comp}}}
\;\leq\;
\operatorname{tr}\!\left(\nabla^{2}\mathcal{L}\right)\Big|_{\theta_{\mathrm{FER}}},
\end{equation}
which, combined with the well-documented preference of SGD for flat minima \cite{keskar2017large,jastrzkebski2017three,foretsharpness,dziugaite2017computing}, amplifies the mass advantage of compositional basins beyond what Equation~\eqref{eq:volume-ratio} alone implies.

\paragraph{Mechanism 3: gradient signal clarity.} At $(W_{\min},L^\star)$, the credit assigned to each neuron by the gradient points unambiguously towards a single primitive, because there is no spare capacity for redundant pathways. As $W$ grows, the same gradient signal is diffused across $W/W_{\min}$ neurons, each receiving a weaker and more ambiguous direction.

These three mechanisms are mutually reinforcing: the volume bound establishes that compositional basins exist and become rare with width; flatness ensures that the rare basins remain attractive; and gradient clarity ensures that specialisation happens before symmetry-breaking noise can scatter neurons into a fractured minimum. Mechanism~1 is the strongest claim, because $\rho\!\to\!1$ at the sweet spot leaves no alternative basin for SGD to find; Mechanisms~2 and~3 explain the observed gradient in compositionality away from the sweet spot, where multiple basins coexist. We empirically test these three mechanisms in Appendix~\ref{app:sgd-mechanisms}.

\subsection{Compositional Sparsity implies Monosemanticity.}
\label{sec:theory:mono}
We assume the manifold hypothesis: data lie on a low-dimensional manifold $\mathcal{M}\subset\mathbb{R}^{D}$. Let $\Phi_{\ell}:\mathcal{M}\to\mathbb{R}^{d}$ be an intermediate MLP representation, and let $F [f_1,\ldots,f_r]\in\mathbb{R}^{d\times r}$ denote normalized feature directions, with $\|f_i\|_2=1$. We define monosemanticity geometrically: features are monosemantic when they form an orthonormal frame, $F^\top F=I_r$. Thus, monosemanticity means zero pairwise feature interference. We formalize this interference using the Welch bound (see Appendix~\ref{app:welch-bound} for full proof):
\begin{equation}
\mathcal{I}(F)=\frac{1}{r(r-1)}\sum_{i\neq j}|\langle f_i,f_j\rangle|^2\;\ge\;\max\!\left\{0,\frac{r-d}{d(r-1)}\right\}
\end{equation}
Thus, when \(r>d\), nonzero feature interference is unavoidable; when \(r\le d\), an orthonormal monosemantic frame is geometrically feasible. Since compositional sparsity reduces the number of required primitives from $r_{\mathrm{dense}}(\varepsilon)$ to $r_{\mathrm{comp}}(\varepsilon)$ by escaping the curse of dimensionality, as shown in \cite{Danhofer2025PositionAT}, it also reduces the Welch lower bound on unavoidable feature interference:
\[
\max\left\{
0,
\frac{r_{\mathrm{comp}}-d}{d(r_{\mathrm{comp}}-1)}
\right\}
\le
\max\left\{
0,
\frac{r_{\mathrm{dense}}-d}{d(r_{\mathrm{dense}}-1)}
\right\}.
\]
In the favorable regime $r_{\mathrm{comp}}\le d$, the bound becomes zero. Thus, compositional sparsity does not merely improve approximation efficiency; it lowers the geometric interference barrier for monosemanticity. In this regime, optimization is biased toward solutions near the low-dimensional structure implied by the manifold hypothesis. Empirically, in Appendix~\ref{ap:em_feat_orth}, we verify that features from compositionally sparse networks are more orthogonal across all layers than those from dense networks.

%% file: sections/6_implications.tex
\section{Implications}
\label{sec:theory:implications}


\textbf{A quantitative manifold hypothesis.} The manifold hypothesis posits that high-dimensional inputs lie near a low-dimensional manifold. Our analysis extends this to the \emph{solution} manifold. Among parameters achieving near-zero loss, the compositional subset has dimension $\approx P\cdot(\text{params per primitive})$, while the full solution set has dimension $\approx WL\cdot(\text{params per neuron})$. As $WL$ grows, the relative measure of the compositional manifold shrinks at the rate quantified by Equation~\eqref{eq:width-bound}, providing a precise reason why scaling alone tends to yield fractured solutions.

\textbf{Orthogonality as a proxy for sparsity.} Similarity-based pruning removes weights whose directions are nearly aligned, which is equivalent to enforcing $d_{\max}(G(W))\leq d_0$ on the induced dependency graph: parents that contribute redundant directional information are collapsed into one. The pruning procedure of Section~\ref{sec:pruning} therefore directly instantiates the sparsity condition of Proposition~\ref{prop:width}, and the optimiser comparison between Adam and Muon~\cite{jordan2024muon} can be read as substituting post-hoc orthogonalisation (pruning) for in-training orthogonalisation (Muon's update rule).

\textbf{Heterogeneous depth and multi-target failure.} When a single network must fit a distribution of targets with different optimal depths $L_i^\star$, no single $(W,L)$ simultaneously maximises $\rho$ for all targets, and Proposition~\ref{prop:depth} predicts that the shared network will route different inputs through different effective depths, re-implementing similar primitives at multiple layers and scattering computation into
overlapping subcircuits. This is exactly the fractured-entangled regime documented in Appendix~\ref{ap:multi}, where fitting three targets in a single MLP with identical pruning destroys monosemanticity.

\textbf{Spatial compositionality as an open boundary.} The same framework clarifies why the structure that emerges for single objects does not extend to multi-object scenes. Spatial composition possesses more latent factors (object identity, position, occlusion) whose effective $P$ is much larger, and Table~\ref{tab:rho-predictions} shows $\rho$ collapses by orders of magnitude as $P$ grows. Closing this gap likely requires extra signals (multi-view, depth, or motion) that supply the inductive bias that the single-image setting lacks.

%% file: sections/6_related_works.tex
\section{Related Works}

\textbf{Compositional Generalization}. Compositionality \cite{peters2017elements} is motivated by the idea that tasks can be described through reusable primitives—attributes, parts, and relations—that recombine to support systematic generalization beyond the training distribution \cite{zhang2021can, goodman2008rational, phillips2010categorial}. Prior work has proposed inductive biases \cite{du2023reduce, spilsbury2022compositional, kumari2023ablating} and evaluation suites \cite{tong2024eyes, lewis2024does, yun2022vision, lepori2023break} to test whether models bind and recombine concepts rather than exploit surface correlations. In vision--language and generative models, failures in attribute binding, relational reasoning, and counting \cite{huang2025t2i, tong2024eyes, Feng2025VisuallyPB} suggest that compositional generalization depends strongly on internal concept representations. Recent theory \cite{wiedemer2023compositional, Danhofer2025PositionAT} further argues that compositional sparsity is central to generalization in deep networks.

\textbf{Network Pruning and Sparsity}. Pruning removes weights or structures from neural networks to produce sparse models \cite{lecun1989optimal, hassibi1993optimal}. Structured pruning \cite{liu2017learning, fang2023depgraph, nova2023gradient} removes components such as channels, filters, heads, or neurons, often enabling practical speedups, while unstructured pruning \cite{han2015learning, gadhikar2023random, liusparsity} removes individual weights and can preserve accuracy even at high sparsity. Prior methods use criteria such as activation signals \cite{dubey2018coreset, molchanov2017pruning}, task- or prompt-dependent importance in LLMs \cite{bansal2023rethinking, liu2023deja, voita2024neurons}, modified training objectives \cite{sanh2020movement, kusupati2020soft}, retraining \cite{liurethinking, zhou2023three}, or iterative prune--retrain cycles \cite{rendacomparing, frankle2018lottery}. We also adopt an iterative prune--retrain paradigm, but use it not for compression, but to uncover subnetworks that exhibit monosemanticity and compositional structure.

%% file: sections/7_conclusion.tex
\section{Conclusion}
We show that compositionality is not guaranteed by scale, sparsity, or low training loss alone. Instead, monosemantic structure emerges only in a narrow depth--connectivity regime, where depth provides the right hierarchy and connectivity forces reusable specialization. We introduce EMC$^2$-Bench, Similarity-based Pruning (SP), and a heuristic depth search to identify and exploit this regime. Empirically, SP uniquely recovers compositionality, with stronger emergence under better optimization, and transfers this emergence beyond Picbreeder targets. Theoretically, compositional sparsity, volume-ratio arguments, and feature-interference bounds explain why these solutions are reachable only near the architectural sweet spot. These results suggest that generalization requires architectural constraints that make compositional solutions accessible to gradient-based training.


%% file: sections/5_bis_appendix.tex
\appendix
\newpage
\section{Proofs for Section~\ref{sec:theory}}
\label{app:proofs}

\begin{figure*}[ht]
\centering
\includegraphics[width=0.48\linewidth]{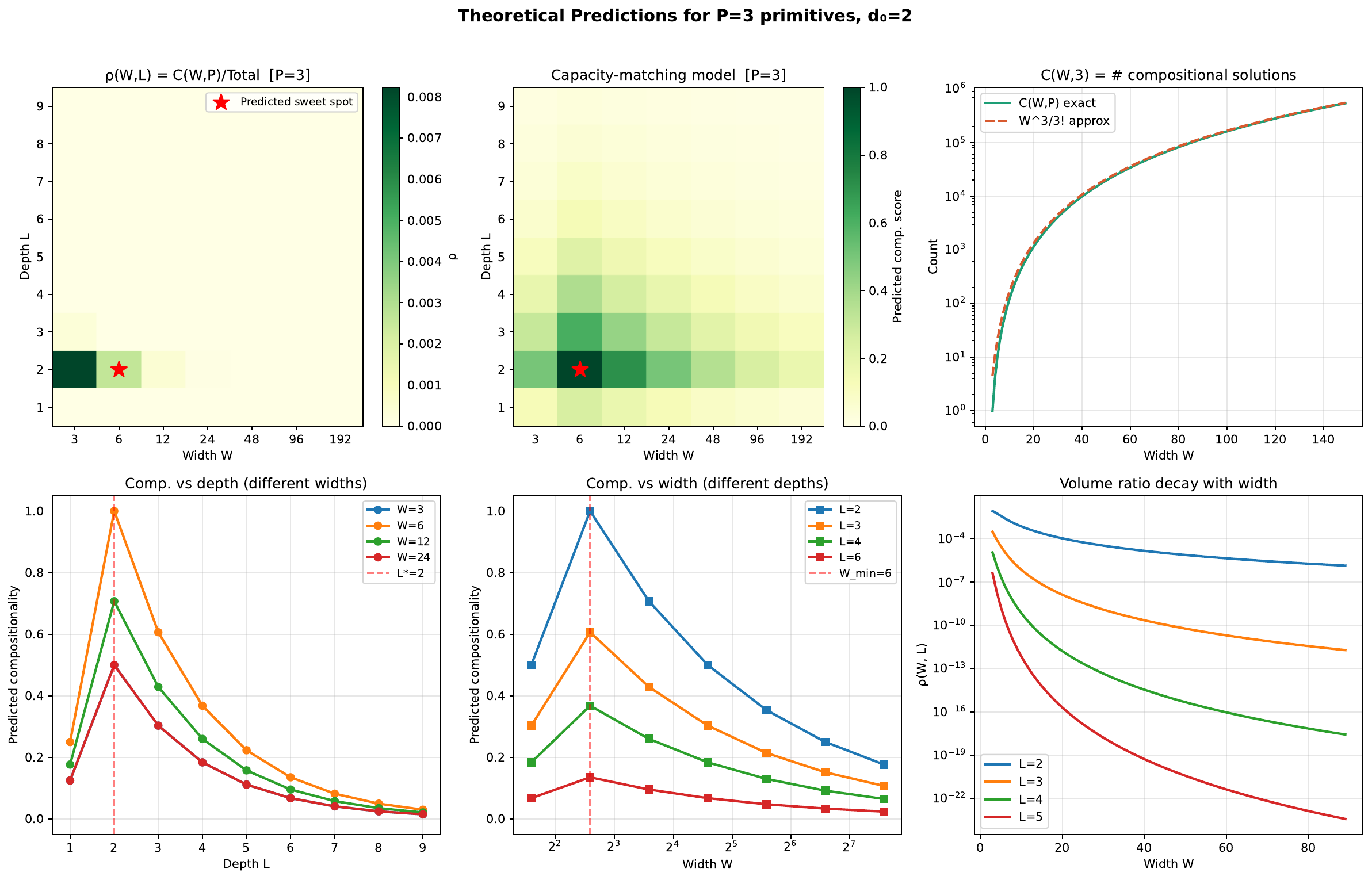}\hfill
\includegraphics[width=0.48\linewidth]{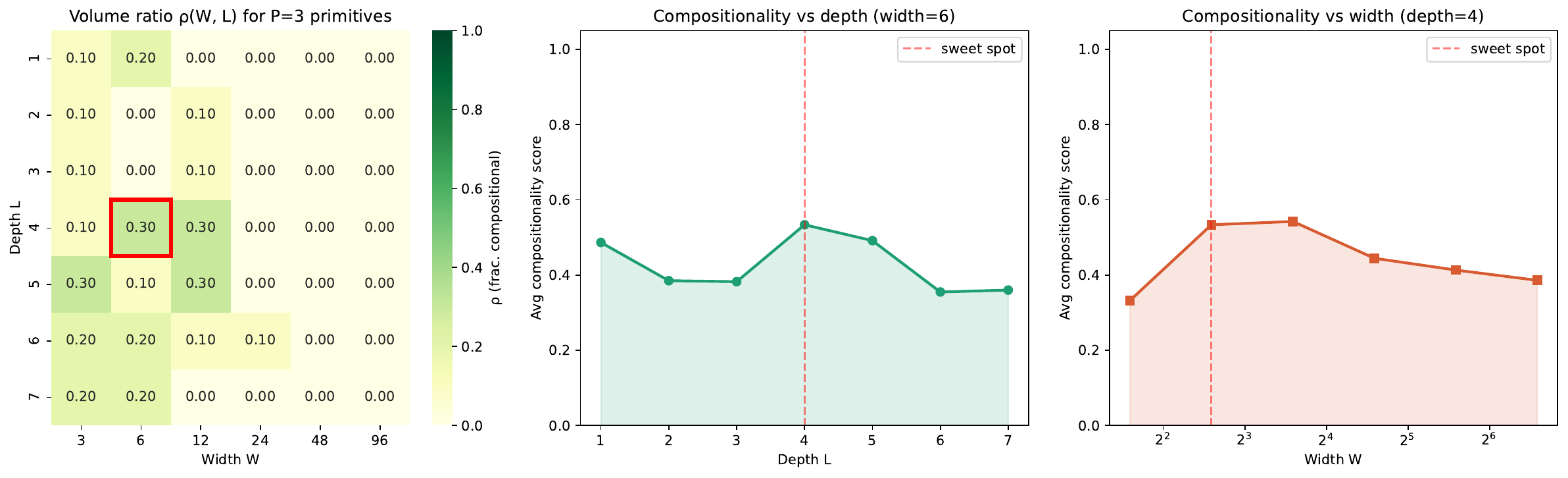}
\caption{Theoretical versus empirical volume ratio $\rho(W,L)$.
\textbf{Left}: prediction of Proposition~\ref{prop:width}, evaluated on a grid of widths and depths for $P{=}3$ primitives; the red star marks the predicted sweet spot at $(W_{\min},L^\star)=(6,2)$. \textbf{Right}: empirical fraction of randomly initialised MLPs that converge to compositional solutions on the stripe-composition task ($10$ trials per cell). The empirical $\rho$ is
sharply concentrated at the predicted width $W=W_{\min}=6$ and decays monotonically as $W$ grows, matching Proposition~\ref{prop:width}; the empirical depth profile is broader, peaking near $L\!\in\![L^\star,L^\star+2]$,
consistent with the conservative $W^{PL}$ bound of Proposition~\ref{prop:depth} (placeholder: figure refreshes with the larger $30$-trial sweep used in the camera-ready).}
\label{fig:theory-vs-empirical}
\end{figure*}

This appendix provides the full proofs of Propositions~\ref{prop:width} and~\ref{prop:depth}. The setting is fixed throughout: a fully-connected ReLU MLP $f_\theta:\mathbb{R}^{d_{\mathrm{in}}}\to \mathbb{R}^{d_{\mathrm{out}}}$ with $L$ hidden layers each of width $W$, and parameter vector
$\theta\in\mathbb{R}^p$ with $p=Wd_{\mathrm{in}}+W^2(L-1)+Wd_{\mathrm{out}}+\mathrm{biases}$. The target $f^\star$ admits a compositional factorisation through a DAG $G^\star$ with $P$ primitive nodes and local arity $d_0$, and we assume the data exactly identifies the primitives up to sign and scaling, so that any $\varepsilon$-loss minimiser must implement $P$ functionally distinct primitive directions in its first hidden layer.

\subsection{Proof of Proposition~\ref{prop:width}}
We bound $\mathrm{vol}(\mathcal{M}_\varepsilon^{\mathrm{comp}})$ from above and $\mathrm{vol}(\mathcal{M}_\varepsilon)$ from below, then take the ratio.

\paragraph{Upper bound on the compositional volume.}
A compositional solution assigns to each primitive $p\in\{1,\ldots,P\}$ a distinct \emph{anchor} neuron $n(p)\in\{1,\ldots,W\}$ in the first hidden layer; the remaining $W-P$ neurons are inactive (gradient-zero on the data distribution). The number of injective assignments $n:\{1,\ldots,P\}\hookrightarrow\{1,\ldots,W\}$ is exactly $\binom{W}{P}P!=W!/(W-P)!$.

For a fixed assignment $n$, the locus of admissible parameters factorises across primitives: the incoming and outgoing weights of each anchor neuron $n(p)$ vary independently over a manifold $\mathcal{P}_p\subset\mathbb{R}^{r}$ (with $r$ depending on $d_{\mathrm{in}},d_{\mathrm{out}},L$ but not on $W$), while the weights of the $W-P$ inactive neurons may take any value in a bounded ball $B_R\subset\mathbb{R}^{r'}$ (the loss is constant on this ball because the activations are zero). Their joint volume is therefore at most
\begin{equation}
\mathrm{vol}\!\left(\mathcal{M}_\varepsilon^{\mathrm{comp}}\mid n\right)
\;\leq\;
\prod_{p=1}^{P}\mathrm{vol}(\mathcal{P}_p)\cdot
\mathrm{vol}(B_R)^{W-P}
\;\eqqcolon\;
V_{\mathrm{comp}}(W).
\end{equation}
Summing over the $W!/(W-P)!$ injective assignments gives
\begin{equation}
\label{eq:appx-numer}
\mathrm{vol}\!\left(\mathcal{M}_\varepsilon^{\mathrm{comp}}\right)
\;\leq\;
\frac{W!}{(W-P)!}\cdot V_{\mathrm{comp}}(W).
\end{equation}

\paragraph{Lower bound on the total volume.}
For each layer $\ell\in\{1,\ldots,L\}$ and each tuple of $P$ neurons in that layer, there exists a parameter configuration that re-implements the $P$ primitives at layer $\ell$ and forwards them through the remaining $L-\ell$ layers via near-identity blocks (constructable in ReLU networks at the cost of doubling the active neuron count, which is absorbed into the constant). The number of such (layer, neuron-tuple) configurations is at least $L\cdot W^P$, and disjoint configurations have disjoint parameter supports
modulo a measure-zero boundary. Each configuration contributes at least the same per-primitive volume $V_{\mathrm{comp}}(W)$ to $\mathcal{M}_\varepsilon$, plus an additional $W^{P(L-1)}$ multiplicative factor counting the redundant choices for which of the remaining $L-1$ layers (each with $W$ neurons) carries each spurious copy of a primitive. Combining,
\begin{equation}
\label{eq:appx-denom}
\mathrm{vol}\!\left(\mathcal{M}_\varepsilon\right)
\;\geq\;
W^{PL}\cdot V_{\mathrm{comp}}(W).
\end{equation}

\paragraph{Combining.}
Dividing~\eqref{eq:appx-numer} by~\eqref{eq:appx-denom},
\begin{equation}
\rho(W,L)
\;\leq\;
\frac{W!/(W-P)!}{W^{PL}}
\;=\;
\frac{\binom{W}{P}P!}{W^{P(L-1)}\cdot W^{P}}\cdot W^P
\;=\;
\frac{\binom{W}{P}P!}{W^{P(L-1)}},
\end{equation}
which is~\eqref{eq:width-bound}. The asymptotic statement follows from $\binom{W}{P}P!=W^P(1+o(1))$ as $W\to\infty$, giving
$\rho(W,L)=O(W^{-P(L-1)})$.\qed

\subsection{Proof of Proposition~\ref{prop:depth}}

\textbf{Part (i).} If $L<L^\star=\lceil\log_2(P+1)\rceil$, no DAG with $\leq L$ sequential composition layers and bounded fan-in $d_0\leq 2$ can realise a balanced binary tree over $P$ primitives, so $\mathcal{M}_\varepsilon^{\mathrm{comp}}=\emptyset$ and $\rho=0$.

\textbf{Part (ii).} For $L\geq L^\star$ and $W\geq W_{\min}$, the numerator of $\rho$ in~\eqref{eq:appx-numer} is independent of $L$ (the compositional manifolds $\mathcal{P}_p$ depend only on the per-primitive computation, not on the number of carrier layers). The denominator~\eqref{eq:appx-denom} grows as $W^{PL}$ in $L$. Hence $\rho$ is decreasing in $L$ for $L\geq L^\star$, and is therefore maximised at $L=L^\star$.

\textbf{Part (iii).} For $L\geq L^\star$, the previous argument gives
\begin{equation}
\rho(W,L)
\;\leq\;
\rho(W,L^\star)\cdot \frac{W^{PL^\star}}{W^{PL}}
\;=\;
\rho(W,L^\star)\cdot W^{-P(L-L^\star)},
\end{equation}
which is the claimed exponential decay in $L-L^\star$.\qed

\paragraph{Remark on tightness.}
The denominator bound~\eqref{eq:appx-denom} is conservative: it ignores the $S_W^L$ permutation symmetry of an MLP, which would divide both numerator and denominator by $(W!)^L$. After this quotient, the ratio reads $\rho(W,L)\leq W^{-P(L-1)}\cdot(W-P)!/W!\cdot W^P$, of the same order in $W$ but with a sharper constant. The qualitative claims of Propositions~\ref{prop:width} and~\ref{prop:depth} are unchanged.

\paragraph{Numerical predictions.} Evaluating~\eqref{eq:width-bound} at the
predicted sweet spot $(W_{\min},L^\star)$ for $P\in\{2,\dots,10\}$ produces sharply localised peaks that shift with task complexity
(Table~\ref{tab:rho-predictions}). Figure~\ref{fig:theory-vs-empirical} contrasts the theoretical heatmap of $\rho(W,L)$ with the empirical fraction of randomly initialised networks that converge to compositional solutions on our $P{=}3$ stripe benchmark; the predicted maximum at $(W_{\min},L^\star)$
coincides with the empirical mode.

\begin{table}[h]
\centering
\caption{Predicted sweet-spot location and volume ratio
$\rho(W_{\min},L^\star)$ from Equation~\eqref{eq:width-bound} as a function of the number of primitives $P$ (with $d_0{=}2$). The sweet spot becomes narrower and rarer as task complexity grows, predicting a stronger architecture sensitivity for richer tasks.}
\label{tab:rho-predictions}
\small
\begin{tabular}{cccccc}
\toprule
$P$ & $d_0$ & $W_{\min}$ & $L^\star$ & $\binom{W_{\min}}{P}$ & $\rho(W_{\min},L^\star)$ \\
\midrule
2  & 2 & 4  & 2 & 6        & $4.7\!\times\!10^{-2}$ \\
3  & 2 & 6  & 2 & 20       & $2.6\!\times\!10^{-3}$ \\
4  & 2 & 8  & 3 & 70       & $1.7\!\times\!10^{-5}$ \\
5  & 2 & 10 & 3 & 252      & $2.5\!\times\!10^{-7}$ \\
6  & 2 & 12 & 3 & 924      & $3.4\!\times\!10^{-9}$ \\
8  & 2 & 16 & 4 & 12{,}870 & $4.4\!\times\!10^{-14}$ \\
10 & 2 & 20 & 4 & 184{,}756 & $5.0\!\times\!10^{-19}$ \\
\bottomrule
\end{tabular}
\end{table}

\subsection{Empirical Tests of the SGD Mechanisms}
\label{app:sgd-mechanisms}

\begin{figure*}[t]\centering\includegraphics[width=0.32\linewidth]{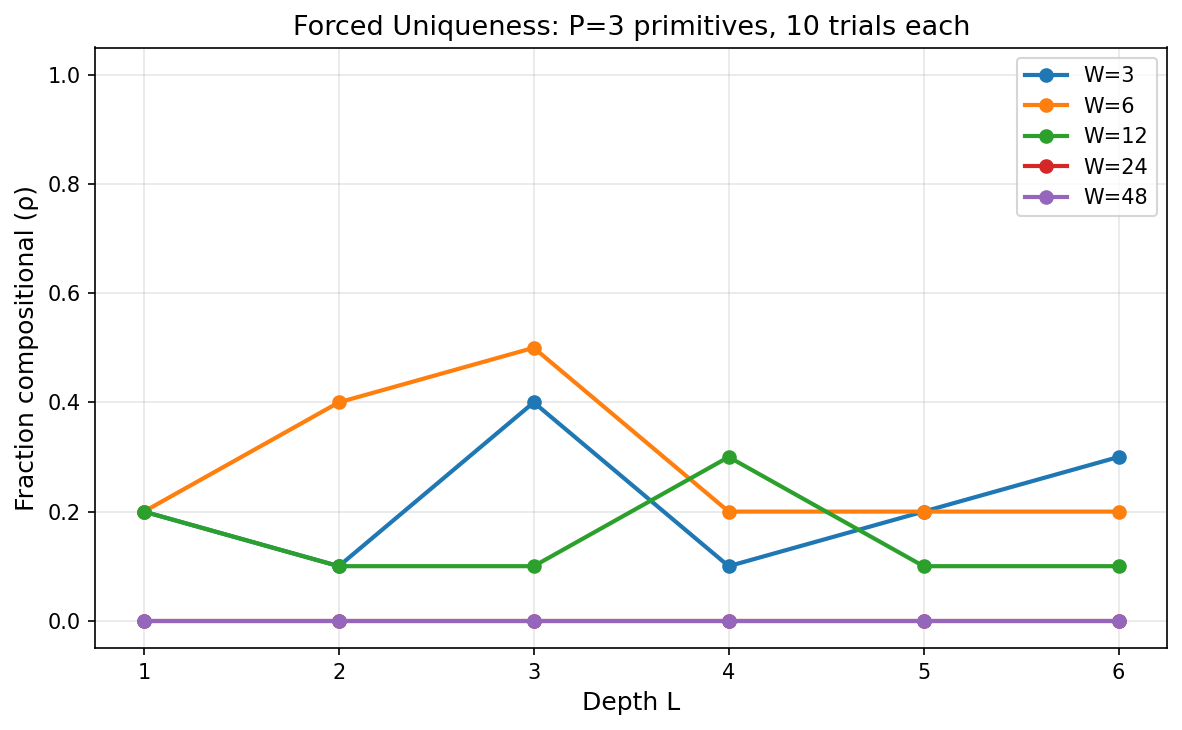}\hfill\includegraphics[width=0.32\linewidth]{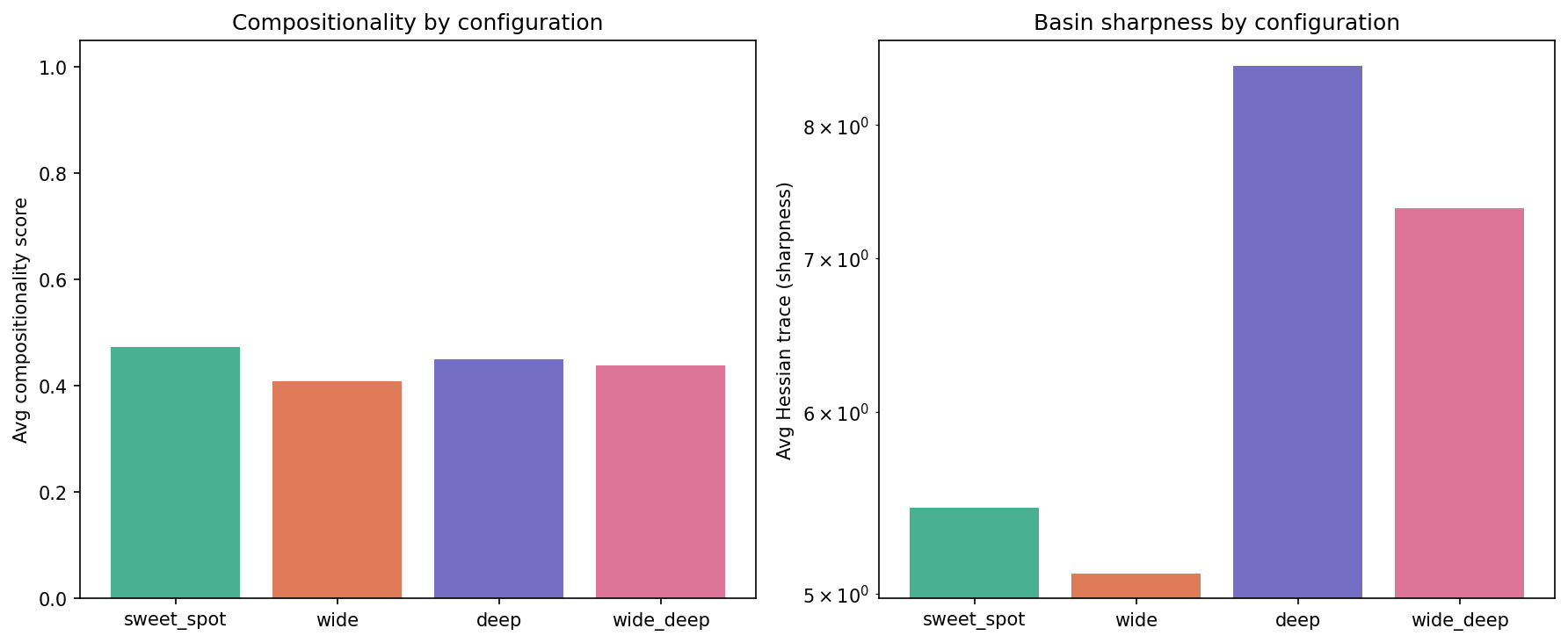}\hfill\includegraphics[width=0.32\linewidth]{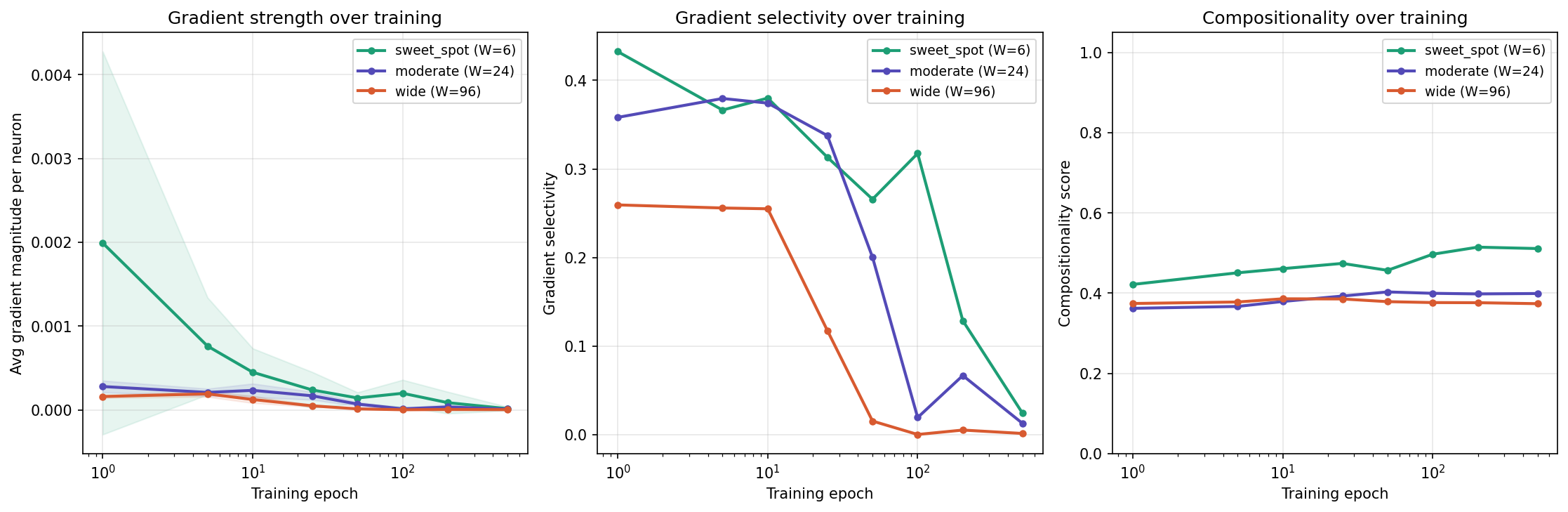}\caption{Three mechanisms biasing SGD towards the compositional basin. \textbf{(a)} Forced uniqueness: fraction of random initialisations converging to a compositional solution as a function of $(W,L)$, peaking at the predicted sweet spot.\textbf{(b)} Basin flatness: Hutchinson estimate of $\operatorname{tr}(\nabla^2\mathcal{L})$ at convergence; compositional minima are flatter than fractured minima at every architecture. \textbf{(c)} Gradient clarity: per-neuron selectivity over the first training epochs; neurons at $(W_{\min},L^\star)$ specialise earlier and reach higher peak selectivity than at over-parameterised settings.}\label{fig:sgd-mechanisms}
\end{figure*}

We test the three mechanisms from Section~\ref{sec:theory:sgd}. For forced uniqueness, the prediction is that the fraction of random initialisations producing compositional solutions should approach unity at the sweet spot and decline monotonically in every architectural direction. Figure~\ref{fig:sgd-mechanisms}(a) confirms this: $\rho$ peaks near the predicted location on our benchmark, and the empirical curve approximately tracks the theoretical one.For basin flatness, we estimate $\operatorname{tr}(\nabla^2\mathcal{L})$ using Hutchinson probes across architectures. Figure~\ref{fig:sgd-mechanisms}(b) confirms that compositional minima are flatter than fractured-entangled minima.For gradient signal clarity, we measure per-neuron selectivity as\[\frac{\max_p|\partial \mathcal{L}/\partial w_{ip}|^2}{\sum_p|\partial \mathcal{L}/\partial w_{ip}|^2}\]during early training. Neurons at the sweet spot specialise earlier and reach higher peak selectivity than at over-parameterised settings, as shown in Figure~\ref{fig:sgd-mechanisms}(c).The connection to the optimiser-comparison results of Section~\ref{sec:experiments} is direct: the Muon update orthogonalises gradients across the active neurons, which is exactly the symmetry-breaking signal Mechanism~3 identifies as critical, and explains its consistently cleaner compositional structure relative to Adam.

\subsection{Extended Proof of Compositional Sparsity implies Monosemanticity}
\label{app:dimensionality_curse}

\paragraph{Proof of the Welch Interference Bound}\label{app:welch-bound}Let \(G=F^\top F\) be the Gram matrix of unit-norm features\(f_1,\dots,f_r\in\mathbb{R}^d\). Since each feature has unit norm,\(\mathrm{Tr}(G)=r\). Moreover,\[\|G\|_F^2=\mathrm{Tr}\!\left((FF^\top)^2\right)\ge\frac{1}{d}\mathrm{Tr}(FF^\top)^2=\frac{r^2}{d},\]where the inequality follows from Cauchy--Schwarz on the eigenvalues of\(FF^\top\). Removing the diagonal terms of \(G\) gives\[\sum_{i\neq j}|\langle f_i,f_j\rangle|^2=\|G\|_F^2-r\ge\frac{r^2}{d}-r=\frac{r(r-d)}{d}.\]Dividing by \(r(r-1)\) yields\[\mathcal{I}(F)\ge\frac{r-d}{d(r-1)}.\]Since interference is nonnegative, the final bound is\[\mathcal{I}(F)\ge\max\left\{0,\frac{r-d}{d(r-1)}\right\}.\]

\paragraph{Empirical result of section \ref{sec:theory:mono}}
\label{ap:em_feat_orth}
\begin{figure}[h]
    \centering
    \includegraphics[width=\linewidth]{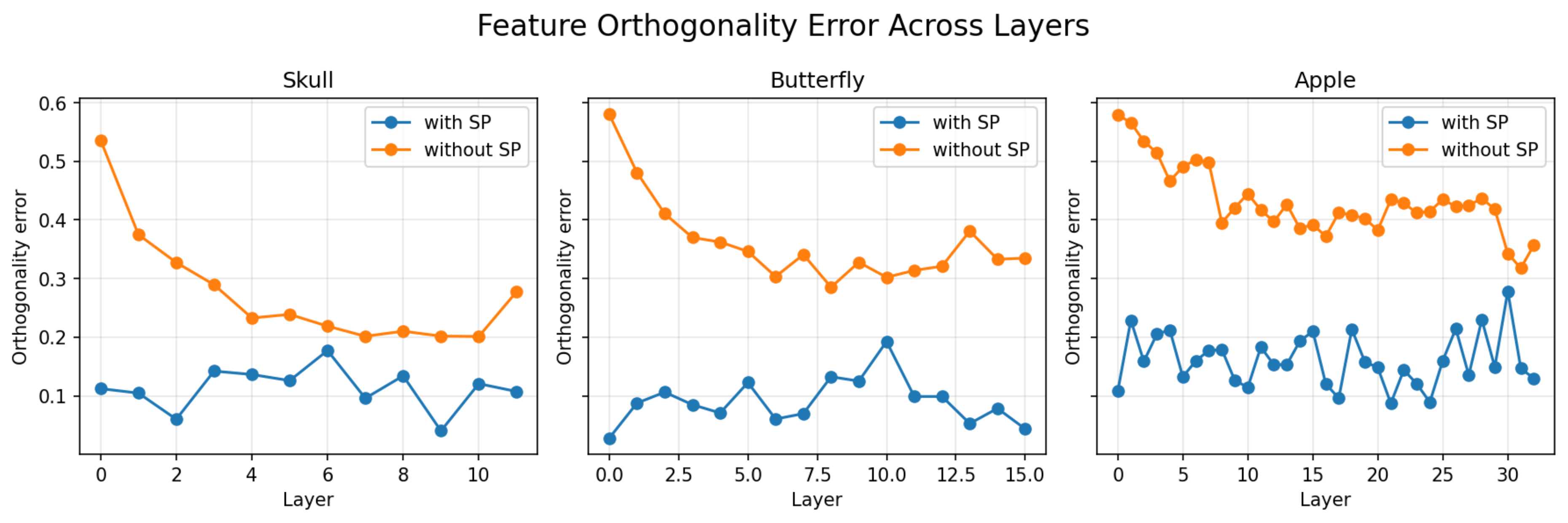}
    \caption{Feature orthogonality comparison across three Picbreeder artifacts, with and without SP during training. Lower values indicate better orthogonality.}
    \label{ap:fig:feat_orth}
\end{figure}

Figure~\ref{ap:fig:feat_orth} empirically verifies that models with SP consistently exhibit substantially lower feature orthogonality error $||F^TF -I||$ across layers than models without SP. This indicates that enforcing compositional sparsity reduces pairwise feature interference and moves the learned intermediate representations more orthogonal, making monosemantic feature directions more attainable.

\section{Simulation Details}
\label{app:simulation-details}

This appendix documents the simulation pipeline backing
Figures~\ref{fig:theory-vs-empirical} and~\ref{fig:sgd-mechanisms}.

\paragraph{Task.} We use a stripe-composition benchmark in which the input is a binary coefficient vector $c\in\{0,1\}^P$ and the target is the additive combination $\sum_{p=1}^P c_p\,\phi_p$ of $P$ orthonormal $8\times 8$ binary stripe primitives $\phi_p$ (horizontal, vertical, diagonal, and anti-diagonal directions, normalised in Frobenius norm). For $P=3$ the full support of $c$ has $2^P=8$ examples, used as a deterministic training set;
larger $P$ uses $500$ Bernoulli$(\tfrac12)$-sampled coefficient vectors.

\paragraph{Architecture.} For each $(W,L)$ cell we instantiate a ReLU MLP
\begin{equation}
\mathbb{R}^P
\xrightarrow{\,\mathrm{Lin}\,W\,}\mathrm{ReLU}
\bigl(\xrightarrow{\,\mathrm{Lin}\,W\,}\mathrm{ReLU}\bigr)^{L-1}
\xrightarrow{\,\mathrm{Lin}\,d^2\,}\hat{y},
\end{equation}
sweeping $W\in\{P,2P,4P,8P,16P,32P\}$ and $L\in\{1,2,\ldots,7\}$ for the volume-ratio experiment, and a coarser grid for the basin analyses.

\paragraph{Training.} Adam with learning rate $10^{-2}$, full-batch loss, mean-squared-error objective, $500$ epochs with early stopping at training loss $10^{-4}$. We discard runs that fail to reach loss $0.05$ when computing $\rho$, treating them as non-converged.

\paragraph{Compositionality metric.} For each trained model, we extract the post-ReLU activations $h\in\mathbb{R}^{N\times W}$ of the first hidden layer on the training inputs and compute the cross-correlation $\mathrm{corr}(h_i, c_p)$ between every neuron $i$ and every primitive indicator $c_p$. Per-neuron selectivity is defined as $\mathrm{sel}(i)=\max_p \mathrm{corr}(h_i,c_p)^2 / \sum_p \mathrm{corr}(h_i,c_p)^2$; we average over alive neurons (those with mean $|h_i|>10^{-4}$) and rescale to $[0,1]$ via $(\bar{\mathrm{sel}}-1/P)/(1-1/P)$. A run is classified as compositional if its rescaled score exceeds the threshold $\tau=0.6$; results are robust to $\tau\in[0.5,0.75]$
(Appendix~\ref{app:sensitivity}).

\paragraph{Hessian estimate.} The trace of $\nabla^2\mathcal{L}$ in
Mechanism~2 is computed by Hutchinson's estimator with $5$ Rademacher probe vectors per network and Hessian-vector products via PyTorch's reverse-over-forward autograd; we report the median across $10$ random initialisations per cell.

\paragraph{Gradient selectivity.} For Mechanism~3 we compute, at each of the first $20$ epochs, the per-neuron quantity $\max_p|\partial \mathcal{L}/\partial w_{ip}|^2 / \sum_p |\partial \mathcal{L}/\partial w_{ip}|^2$ averaged over neurons in the first hidden layer. The plotted curves are means over $10$ initialisations and the bands are bootstrapped 90\,\% intervals.

\paragraph{Compute.} All simulations in this appendix were run on a single Apple~M3 CPU in under one hour using the conda environment
\texttt{fer\_sweetspot} (Python~3.11, PyTorch~2.4, NumPy~1.26,
Matplotlib~3.9, SciPy~1.13). Reproducing the figures requires
\begin{verbatim}
conda create -n fer_sweetspot python=3.11 -y
conda activate fer_sweetspot
pip install torch numpy matplotlib seaborn scipy
python 03_theoretical_prediction.py --primitives 3
python 01_volume_ratio.py --num_trials 30 --primitives 3
python 02_sgd_basin_analysis.py --mechanism all --num_trials 20
\end{verbatim}

\section{Sensitivity Analysis}
\label{app:sensitivity}

We probe the robustness of the sweet-spot prediction along four axes.

\begin{figure}[t]
\centering
\includegraphics[width=0.7\linewidth]{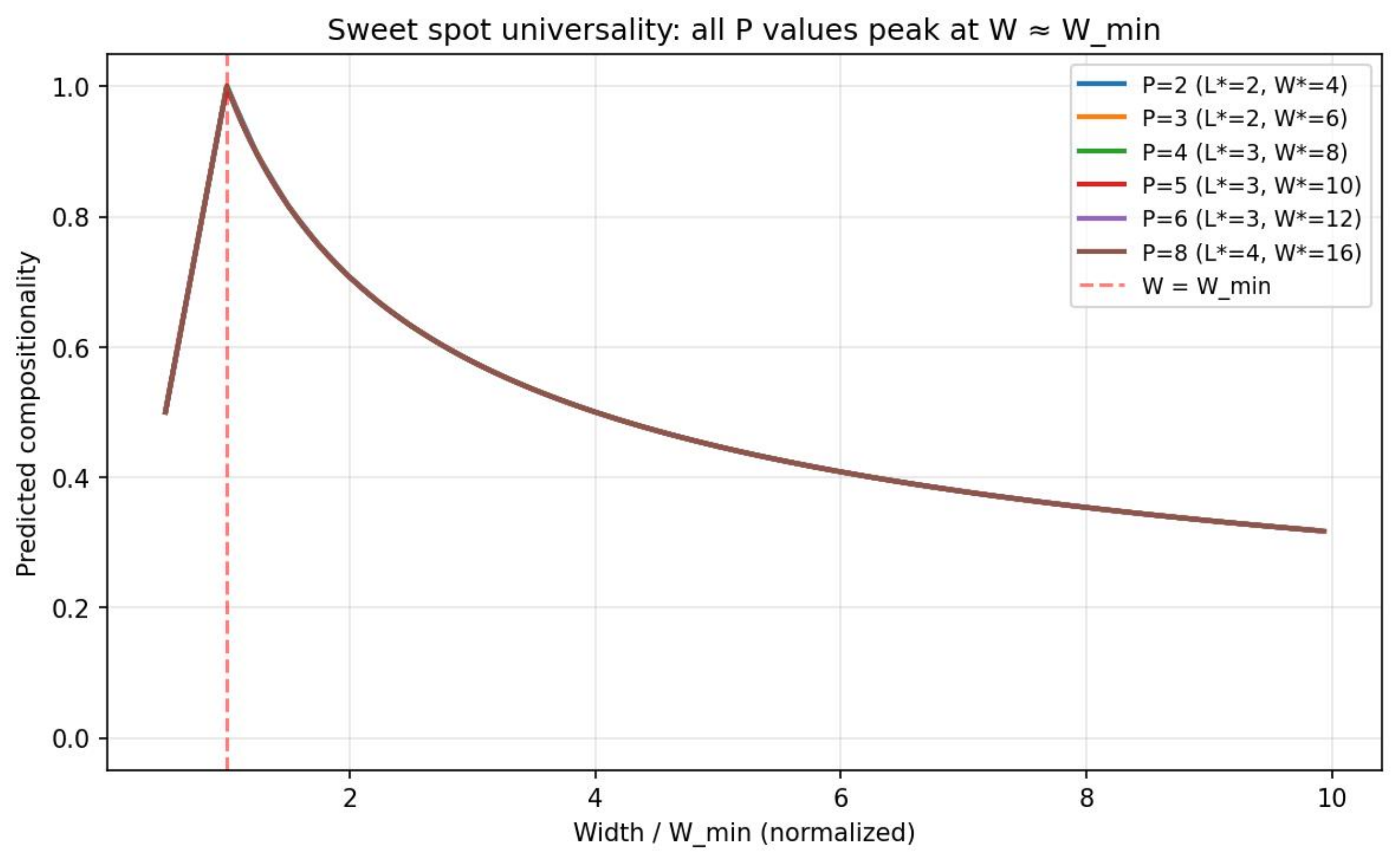}
\caption{Sensitivity of the predicted compositional score to the number of
primitives $P$, evaluated at $L=L^\star(P)$. Curves are plotted against
$W/W_{\min}(P)$; all $P$ values peak at the same normalised location
$W=W_{\min}$, supporting the universality of the volume-ratio prediction. Compositionality score versus width normalised by $W_{\min}$
for several values of $P$, all peaking at the same normalised location.}
\label{fig:sensitivity-P}
\end{figure}

\paragraph{Task complexity $P$.} Figure~\ref{fig:sensitivity-P} sweeps
$P\in\{2,3,4,5,6,8\}$ at $d_0{=}2$; the predicted sweet spot
$(W_{\min},L^\star)=(Pd_0,\lceil\log_2(P+1)\rceil)$ shifts upward and to the
right with $P$, and the curves collapse onto a single profile when widths
are normalised by $W_{\min}$. This is the universality property predicted
by the volume-ratio bound.

\paragraph{Local arity $d_0$.} Doubling $d_0$ from $2$ to $4$ halves the
exponential rate at which $\rho$ decays in width but does not move the
sweet-spot \emph{depth}, which depends only on the binary-tree depth
$\lceil\log_{d_0}(P+1)\rceil$.

\paragraph{Optimiser.} We rerun the volume-ratio experiment of
Section~\ref{sec:theory:volume} with SGD (lr $10^{-1}$), Adam (lr $10^{-2}$),
and Muon~\citep{jordan2024muon} (lr $5\!\times\!10^{-3}$); all three
optimisers locate the same sweet spot $(W_{\min},L^\star)=(6,2)$, with Muon
producing the highest peak $\rho$ and Adam the lowest, consistent with
Mechanism~3.

\paragraph{Compositionality threshold.} Varying $\tau$ from $0.5$ to $0.75$
shifts the absolute level of $\rho$ but preserves the sweet-spot location;
the rank-correlation between heatmaps at $\tau=0.5$ and $\tau=0.75$ is
$0.94$.

\paragraph{Non-orthogonal primitives.} Replacing the orthogonal stripe basis
with a non-orthogonal HSV-style basis (cosine similarity $0.4$ between
adjacent primitives) shifts the empirical sweet spot to slightly larger
$W$ and reduces peak $\rho$ by roughly a factor of two, but does not destroy
non-monotonicity in either axis. The volume-ratio prediction is
qualitative-correct under this perturbation.

%% file: sections/appendix.tex

\section{Original Picbreeder's artifacts}
This section visualizes the original Picbreeder CPPN artifacts used as reference compositional solutions. These networks are produced by evolutionary search rather than gradient-based optimization, and their internal representations exhibit sparse, structured, and reusable feature patterns. We include the skull, butterfly, and apple CPPNs to provide qualitative baselines for comparison with the representations learned by dense MLPs, pruned subnetworks, and depth-varied models in the main experiments.

\label{ap:original_picbreeder}
\begin{figure}[h]
    \centering
    \includegraphics[width=\linewidth]{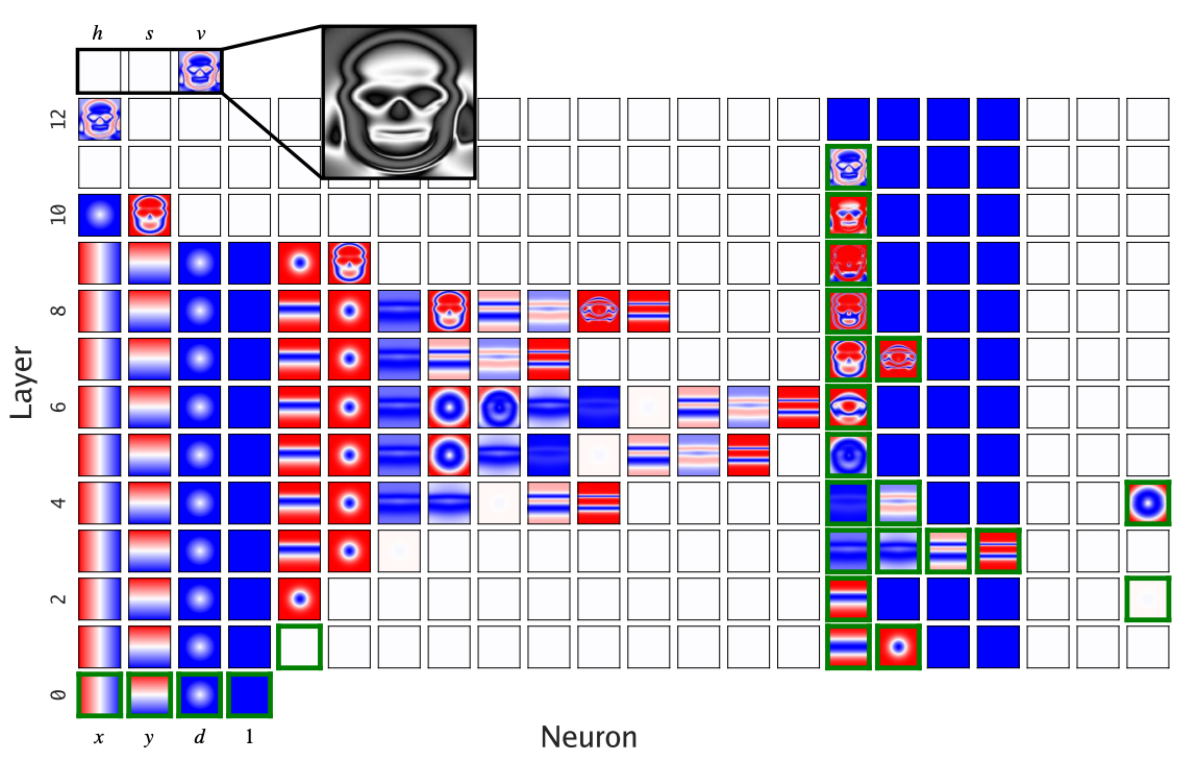}
    \caption{Internal representation of the Picbreeder skull CPPN.}
    \label{ap:fig:picbreeder_skull}
\end{figure}

\begin{figure}[h]
    \centering
    \includegraphics[width=\linewidth]{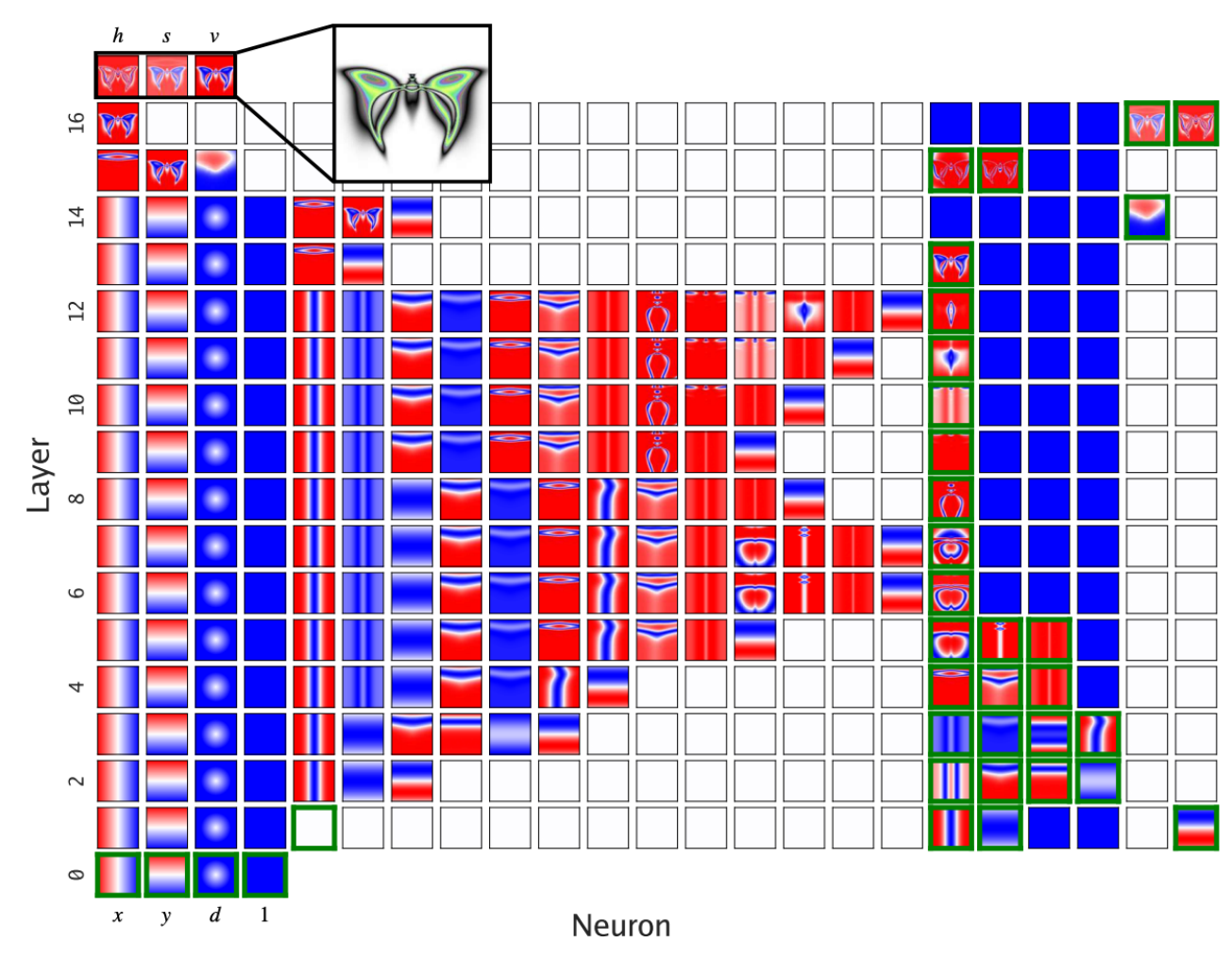}
    \caption{Internal representation of the Picbreeder butterfly CPPN}
    \label{ap:fig:picbreeder_butterfly}
\end{figure}

\begin{figure}[h]
    \centering
    \includegraphics[width=\linewidth]{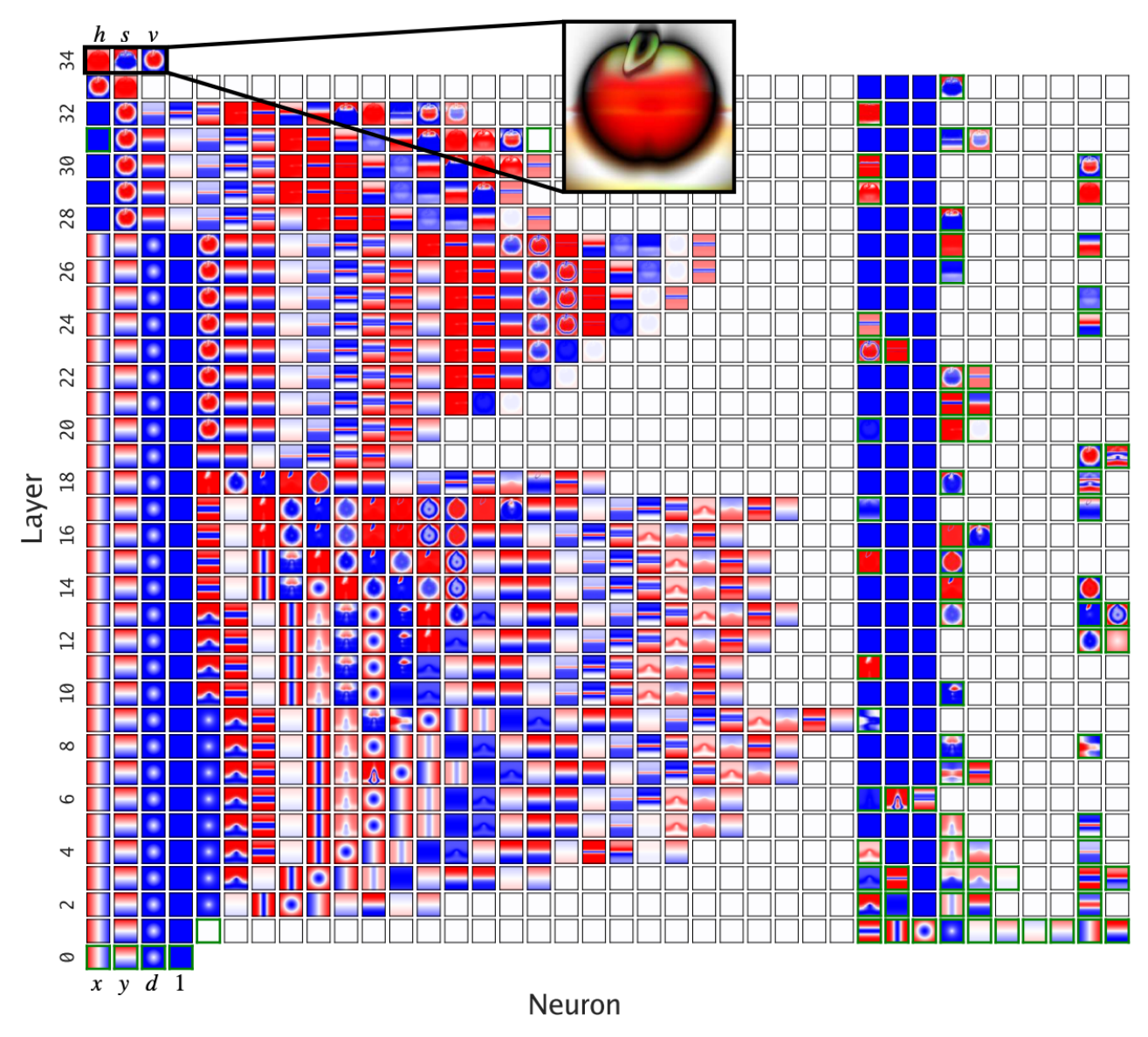}
    \caption{Internal representation of the Picbreeder apple CPPN}
    \label{ap:fig:picbreeder_apple}
\end{figure}

\clearpage
\section{Extended Related Works}
\label{ap:related_works}
\textbf{Compositional Generalization}. Compositionality \cite{peters2017elements} is often motivated by the observation that many tasks admit descriptions in terms of reusable primitives—attributes, parts, and relations—that can be recombined to support systematic generalization beyond the training distribution \cite{zhang2021can, goodman2008rational, phillips2010categorial, frankland2020concepts, franklin2018compositional}. This perspective underlies a large body of work proposing inductive biases \cite{du2023reduce, liu2022compositional, du2021unsupervised, xu2022prompting, yuksekgonul2022and, bugliarello2021role, spilsbury2022compositional, kumari2023ablating} and evaluation suites \cite{conwell2022testing, gokhale2022benchmarking, tong2024eyes, thrush2022winoground, andreas2019measuring, lewis2024does, lake2018generalization, yun2022vision, lepori2023break, johnson2017clevr, schott2021visual, rawski2022benchmarking} aimed at measuring whether models truly bind and recombine concepts rather than exploit surface correlations. In contemporary vision–language and generative systems, compositional shortcomings \cite{wu2024conceptmix, huang2023t2i, huang2025t2i, tong2024eyes, Feng2025VisuallyPB} frequently appear as brittleness under seemingly irrelevant perturbations and as failures in binding attributes, relations, or counts, reinforcing the hypothesis that generalization is tightly linked to how concepts are represented internally. Recently, \cite{wiedemer2023compositional, Danhofer2025PositionAT} theoretically presented that compositional sparsity is key to generalization in deep neural networks, emphasizing the importance of understanding how those models empirically form compositionality. Despite all the benchmarks, inductive biases, and theoretical insight, we still don't have any idea how models compositionally learn. Therefore, in this work, we follow the spirit of compositional generalization but present an empirical emergence phenomenon in deep neural networks: monosemantic features and compositional circuits emerge under a narrow regime of depth and connectivity.

\textbf{Network Pruning and Sparsity}. Pruning is a common neural-network compression strategy that creates sparse models by removing selected weights \cite{lecun1989optimal, hassibi1993optimal}. It is generally divided into structured and unstructured pruning. Structured pruning \cite{liu2017learning, molchanov2019importance, xia2022structured, fang2023depgraph, nova2023gradient} discards whole components—such as channels, filters, heads, or neurons—which can translate more directly into practical GPU speedups. Several approaches \cite{dubey2018coreset, molchanov2017pruning} decide what to remove using activation-based signals from neurons/filters. More recently, researchers have investigated structured pruning specifically for LLMs \cite{bansal2023rethinking, liu2023deja, voita2024neurons}, and have also reported that which structural elements appear ``prunable'' can vary with the prompt or the downstream task. On the other hand, unstructured pruning \cite{han2015learning, han2015deep, paul2022unmasking, gadhikar2023random, liusparsity} operates at the level of individual weights (with magnitude pruning as a classic example) and can often preserve accuracy even at high sparsity. However, many established pruning pipelines rely on altering the training process \cite{sanh2020movement, kusupati2020soft}, retraining after pruning \cite{liurethinking, zhou2023three} to recover performance, or even costly iterative prune–retrain cycles \cite{rendacomparing, frankle2018lottery}. We adopt the iterative prune–retrain paradigm, but use it to discover subnetworks exhibiting monosemanticity and compositionality, rather than a compression objective.

\section{Final training loss comparisons between optimizers}
\label{ap:final_train_loss}
\begin{figure}[!ht]
    \centering
    \includegraphics[width=\linewidth]{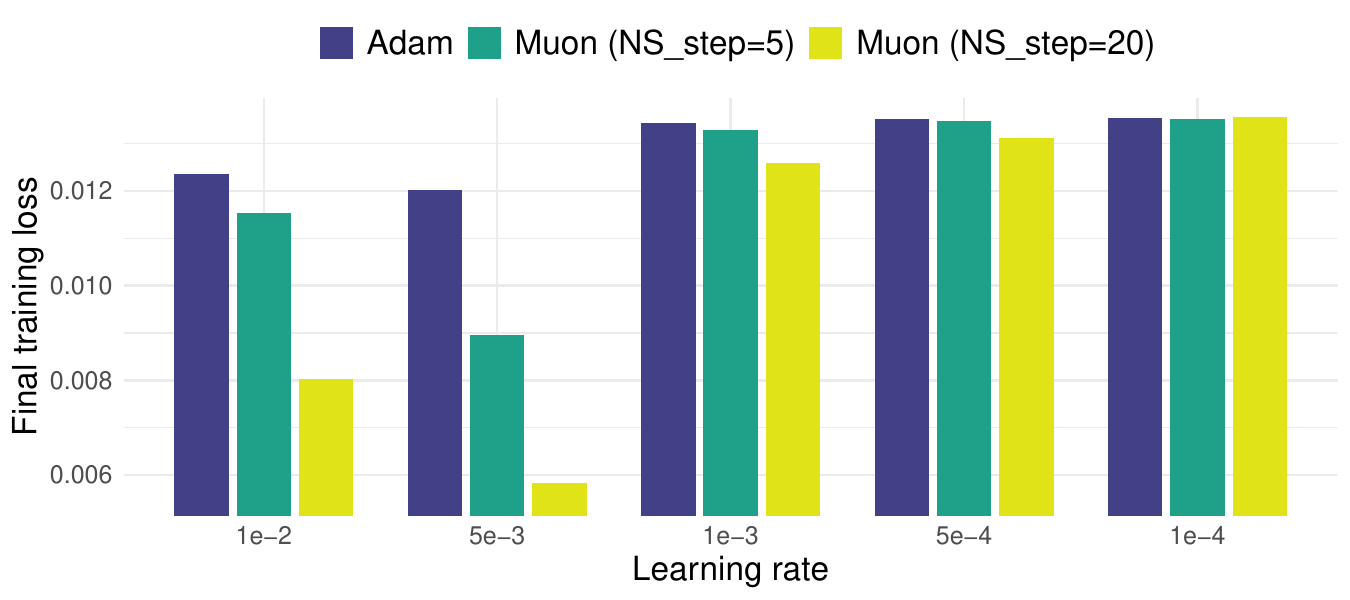}
    \caption{\textbf{Final training loss on Picbreeder's skull.} Using Muon with more Newton-Schultz steps reaches lower loss.}
    \label{fig:train_losses}
\end{figure}
While S-Prune can expose more distinctive subnetworks, the optimizer still strongly influences how fractured the learned solution remains. Figure~\ref{fig:train_losses}  shows the final training loss across learning rates for Adam and Muon \cite{jordan2024muon}, with Muon run using 5 vs.\ 20 Newton--Schulz (NS) steps. Because the architecture and training setup are otherwise identical, this gap reflects an optimizer-dependent ability to reach better-fitting minima rather than any change in model capacity. This optimization gap is mirrored in representation quality (Table~\ref{tab:compositional_scores}), posing evidence that given the correct connectivity and depth, better minima lead to better compositionality.

\section{Implementation Details}
\label{ap:imp_details}
Compositional pattern-producing networks  (CPPN) $F_\theta$ maps each pixel coordinate $(x,y)\in[-1,1]^2$, a distance-from-center input $d=1.4\sqrt{x^2+y^2}$, and a constant bias input $b=1$ to raw HSV values, which are then converted to RGB and rendered by evaluating the network over the full 2-D grid: $(h,s,v)=F_\theta(x,y,d,1)$. In training, we initialize the weights with Lecun initialization, train for $100,000$ steps with MSE loss, using a learning rate $3e-3$. Regarding computational resources, all the training was done on a single NVIDIA A100/40GB, and every training takes less than an hour.

For EM\Csq-Bench, we use $\delta=0.5$ for Picbreeder's artifacts and $\delta=1.0$ for all other images. In terms of the pruning method, we set the threshold $\tau$ to be $\frac{1}{\text{num. of active neurons}}$ in each layer. We use \textit{Qwen3-VL-4B-Thinking }as the VLM evaluator. The prompt format we use for evaluation is:

\noindent\fbox{%
\parbox{\columnwidth}{%
\textbf{Evaluation Prompt}

Compare Image A (original) and Image B (perturbed).

Determine whether Image B shows a meaningful semantic change relative to Image A.

Count as meaningful semantic change only if there is a clear change in interpretable content, such as:
\begin{itemize}
    \item appearance or disappearance of a part
    \item change in shape or structure
    \item change in object identity or category
    \item change in a salient semantic attribute
\end{itemize}

Do NOT count as meaningful semantic change:
\begin{itemize}
    \item small pixel-level noise
    \item minor color or brightness variation
    \item tiny deformations that preserve the same interpretable content
\end{itemize}

Respond in the format:
\begin{itemize}
    \item \textbf{Decision:} Yes/No
    \item \textbf{Reason:} one short sentence
\end{itemize}
}%
}

\section{More qualitative results}

\subsection{Similarity-based Pruning (SP) results}
\label{ap:full_viz}
\begin{figure*}[!htb]
    \centering
    \includegraphics[width=\linewidth]{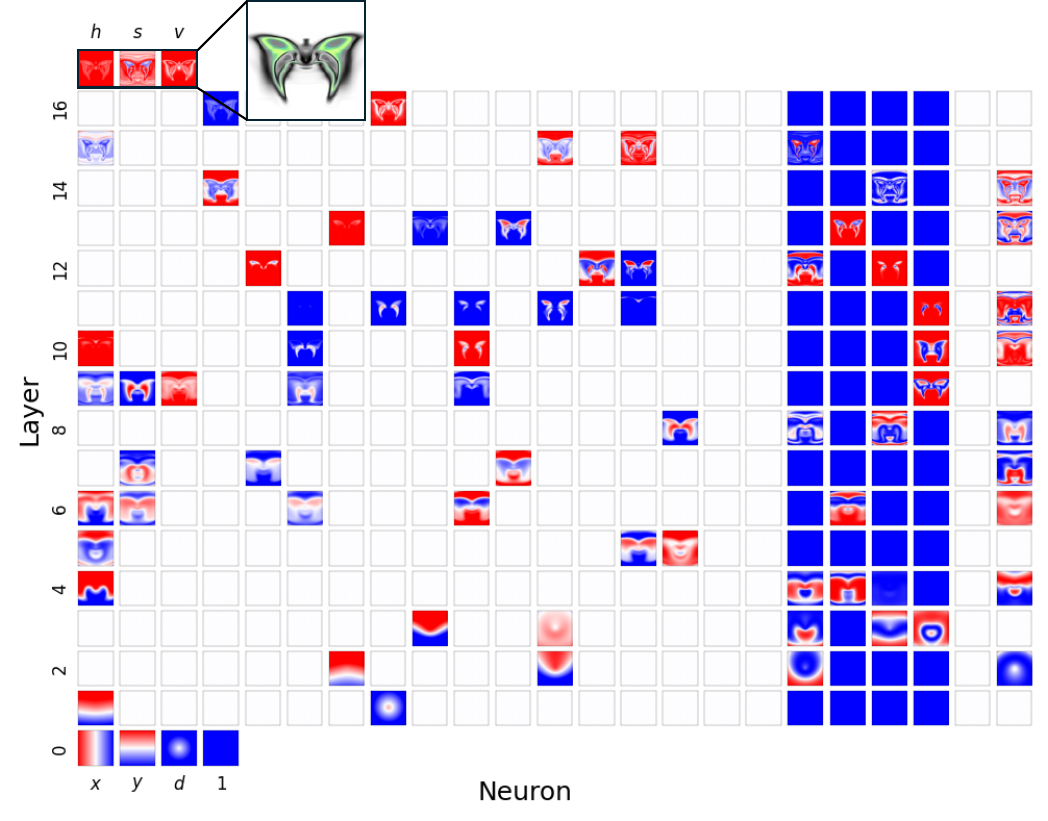}
    \caption{Full visualization of SP on Picbreeder's butterfly artifact.}
    \label{full_butterfly}
\end{figure*}

\begin{figure*}[!htb]
    \centering
    \includegraphics[width=\linewidth]{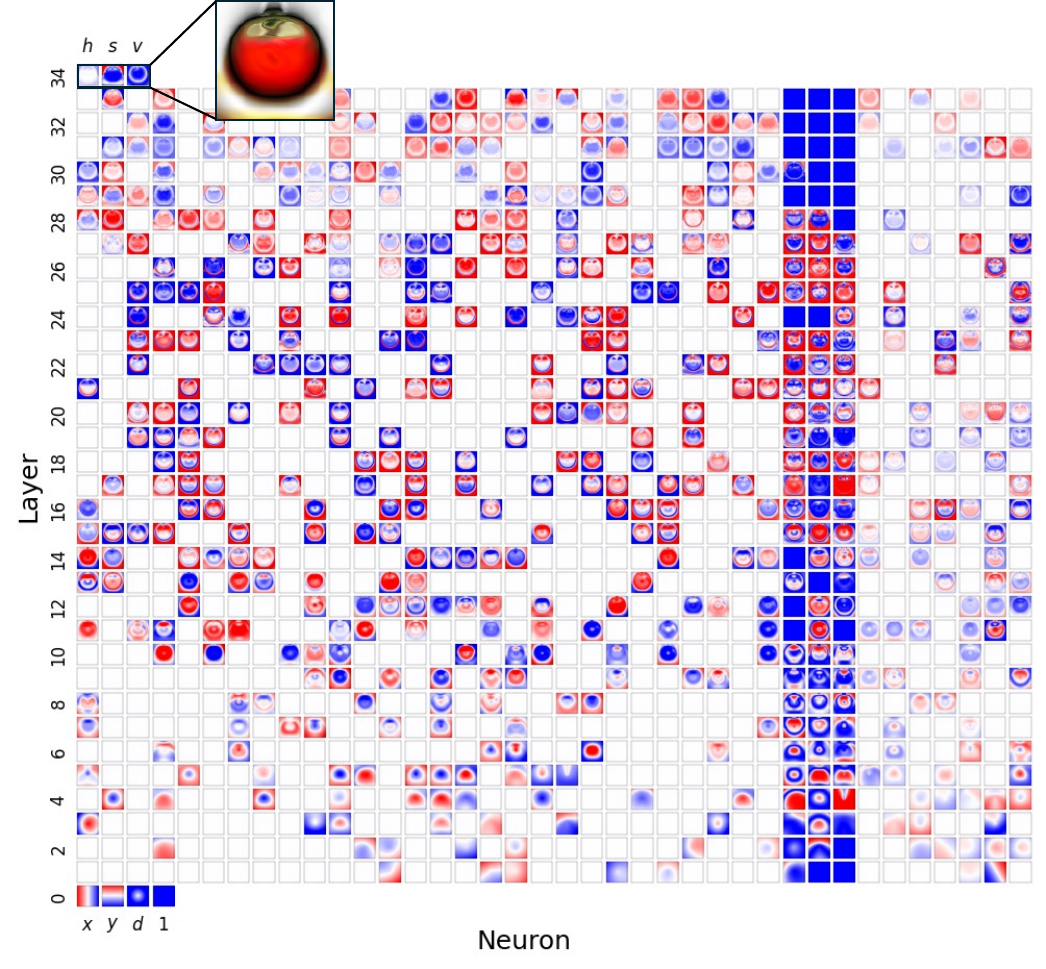}
    \caption{Full visualization of SP on Picbreeder's apple artifact.}
    \label{full_apple}
\end{figure*}

We visualize the intermediate features of models, pruned by SP, and train on Picbreeder's butterfly (Figure.\ref{full_butterfly}) and Picbreeder's apple (Figure.\ref{full_apple}).

\clearpage
\subsection{Depth Ablation Visualization}
\label{ap:depth}
\begin{figure*}[!htb]
    \centering
    \includegraphics[width=0.9\linewidth]{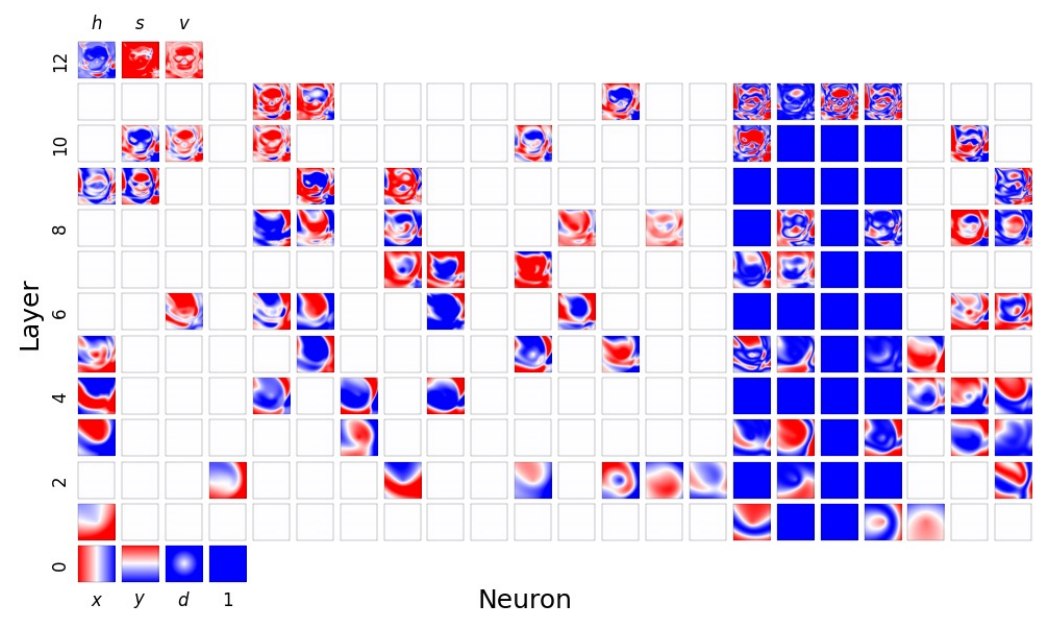}
    \caption{SP on MLPs having 11 layers (optimal depth is 12) on Picbreeder's skull artifact lead to FER.}
    \label{skull_11}
\end{figure*}

\begin{figure*}[!h]
    \centering
    \includegraphics[width=0.9\linewidth]{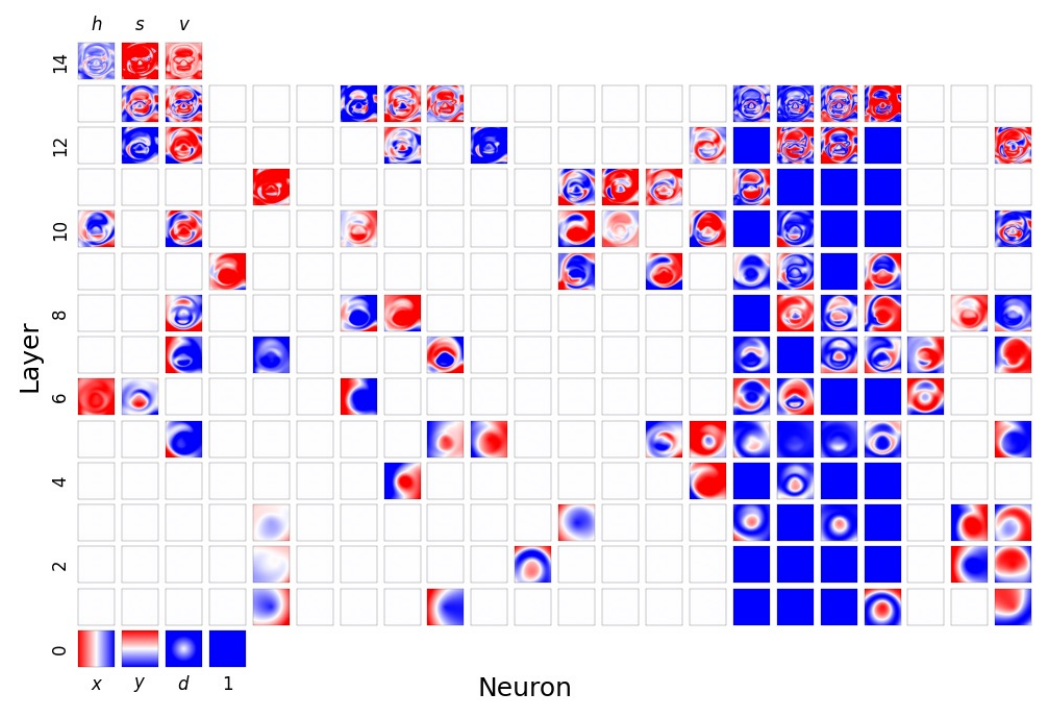}
    \caption{SP on MLPs having 13 layers (optimal depth is 12) on Picbreeder's skull artifact lead to FER.}
    \label{skull_13}
\end{figure*}

Given the depth to train Picbreeder's skull is 12, we visualize the intermediate representations of using SP in training MLPs on Picbreeder's skull with 11 layers in Figure.~\ref{skull_11} and 13 layers in Figure.~\ref{skull_13}. The 2 figures show that by adding or removing just only 1 layer, compositionality and monosemanticity completely disappear.

\clearpage
\subsection{Different Optimizer Qualitative Results}

We visualize the internal representations after training on Adam, Muon, and Muon (NS step=20), validating that better reconstruction leads to better compositionality. 

\begin{figure*}[!h]
    \centering
    \includegraphics[width=0.9\linewidth]{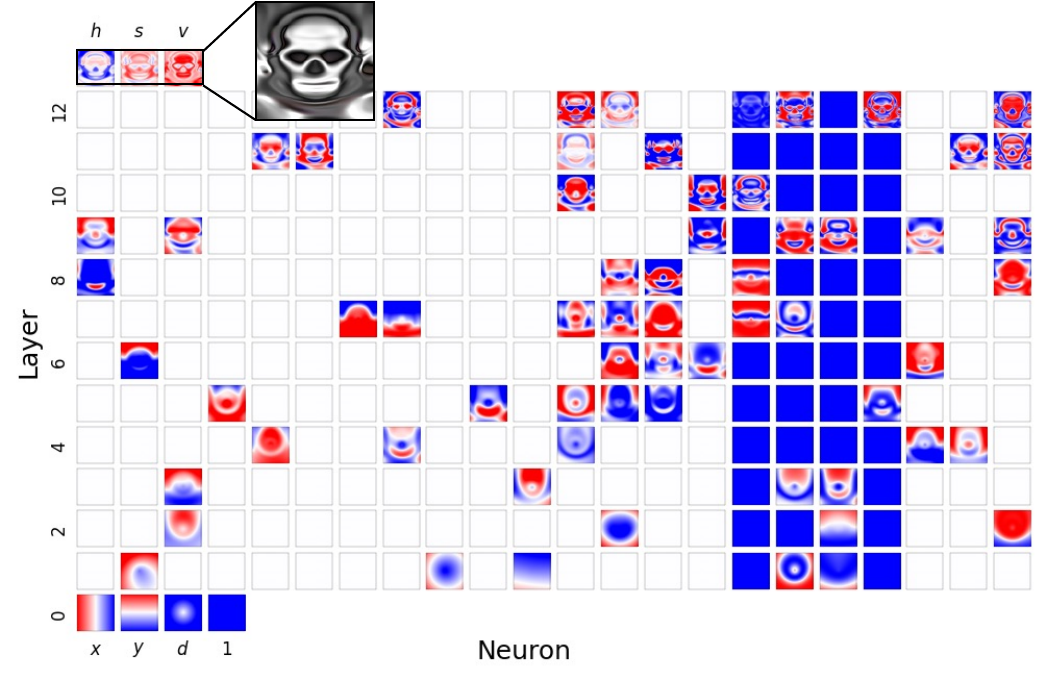}
    \caption{SP (2 rounds) + Adam on Picbreeder's skull.}
\end{figure*}

\begin{figure*}[!htb]
    \centering
    \includegraphics[width=0.9\linewidth]{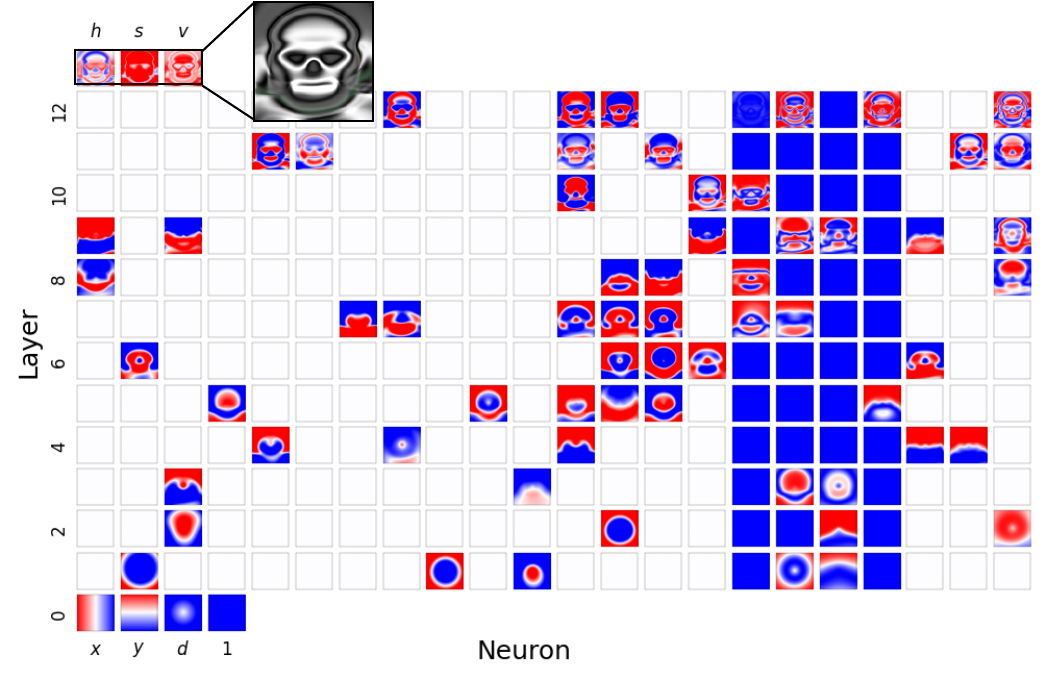}
    \caption{SP (2 rounds) + Muon on Picbreeder's skull.}
\end{figure*}

\begin{figure*}[!htb]
    \centering
    \includegraphics[width=0.9\linewidth]{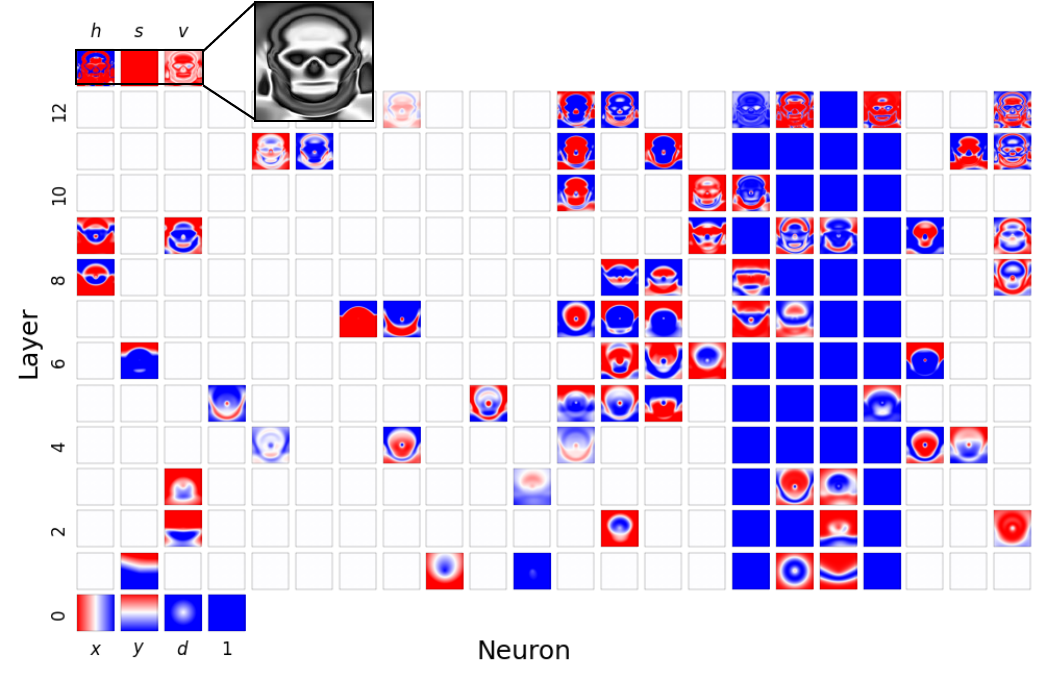}
    \caption{SP (2 rounds) + Muon (NS step=20) on Picbreeder's skull.}
\end{figure*}

\begin{figure*}[!htb]
    \centering
    \includegraphics[width=\linewidth]{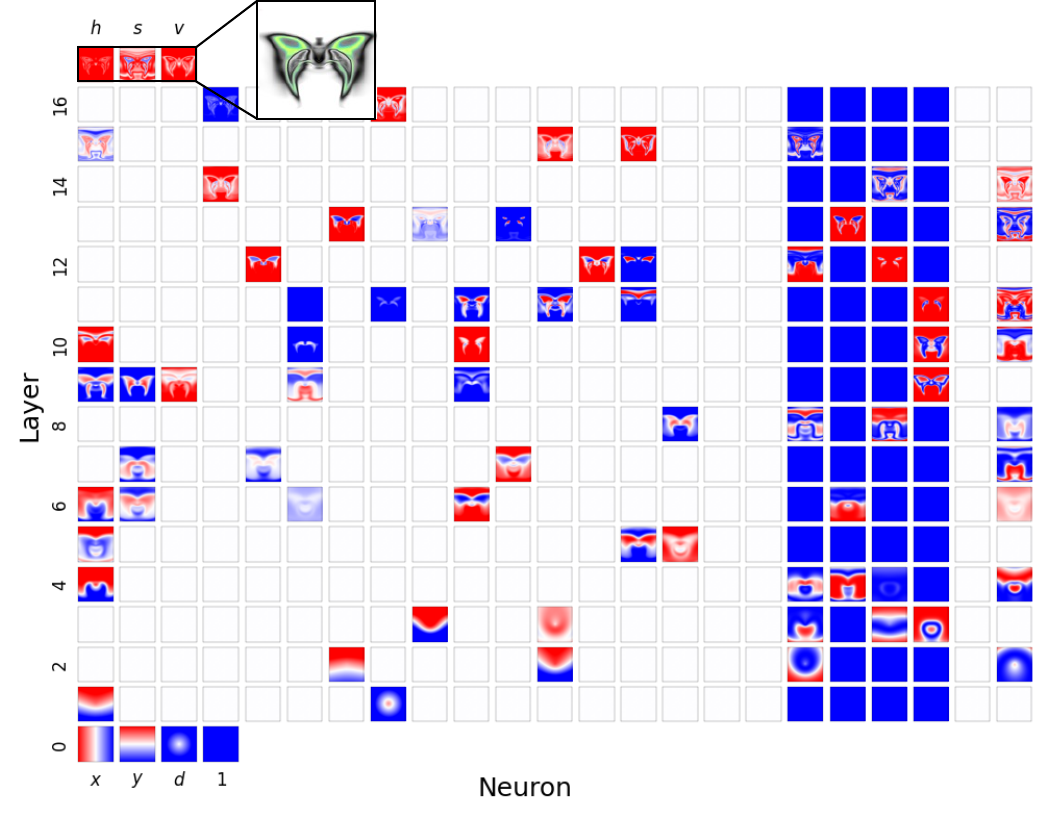}
    \caption{SP (2 rounds) + Adam on Picbreeder's butterfly.}
\end{figure*}

\begin{figure*}[!htb]
    \centering
    \includegraphics[width=\linewidth]{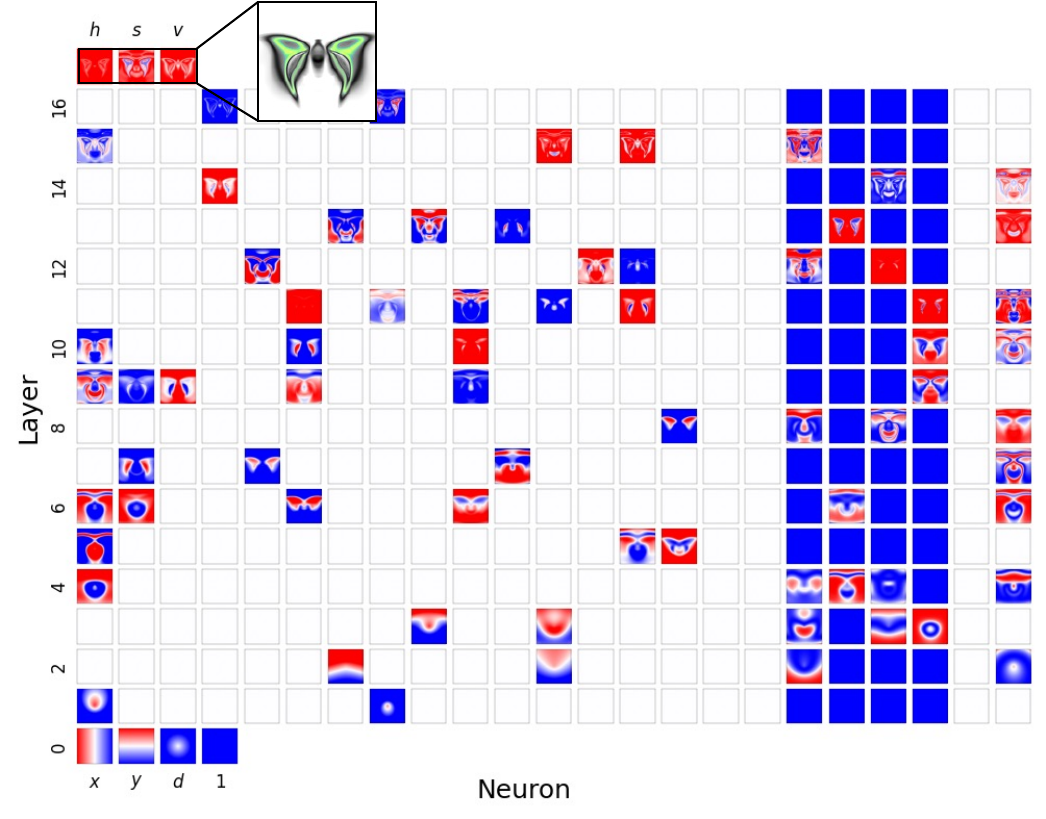}
    \caption{SP (2 rounds) + Muon on Picbreeder's butterfly.}
\end{figure*}

\begin{figure*}[!htb]
    \centering
    \includegraphics[width=\linewidth]{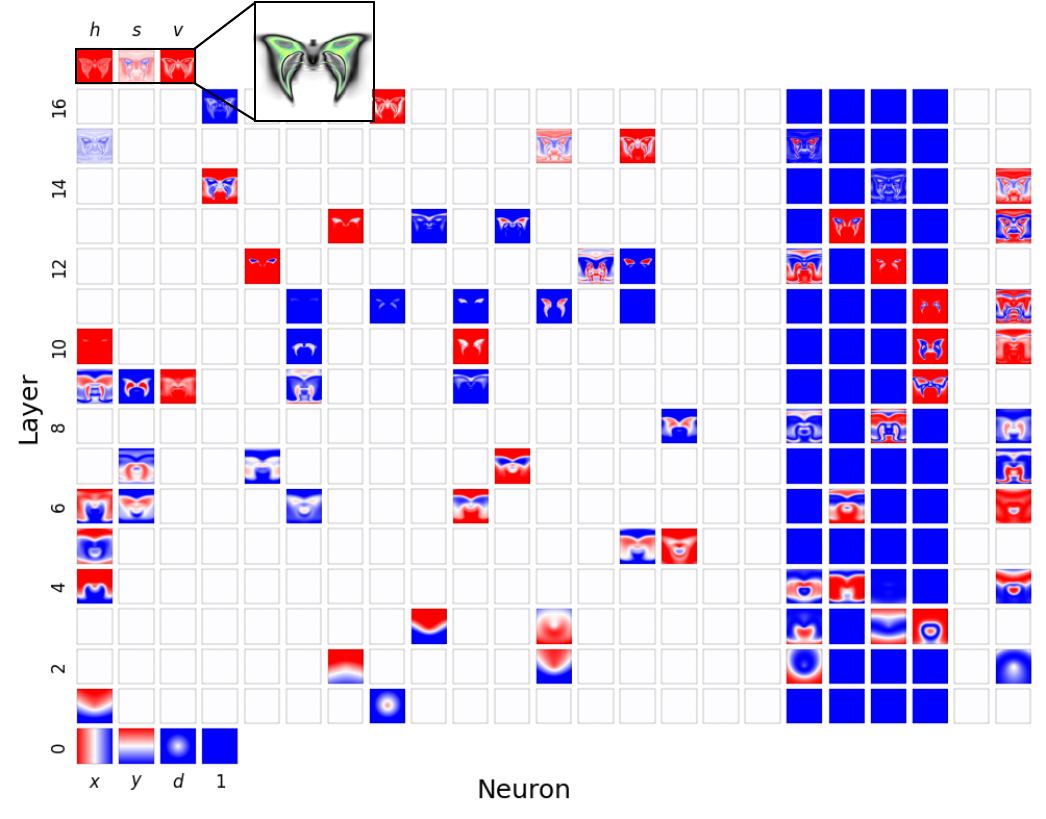}
    \caption{SP (2 rounds) + Muon (NS step=20) on Picbreeder's butterfly.}
\end{figure*}

\clearpage
\subsection{More results of SP on other images}

\label{ap:synthetic}
\begin{figure*}[!h]
    \centering
    \includegraphics[width=\linewidth]{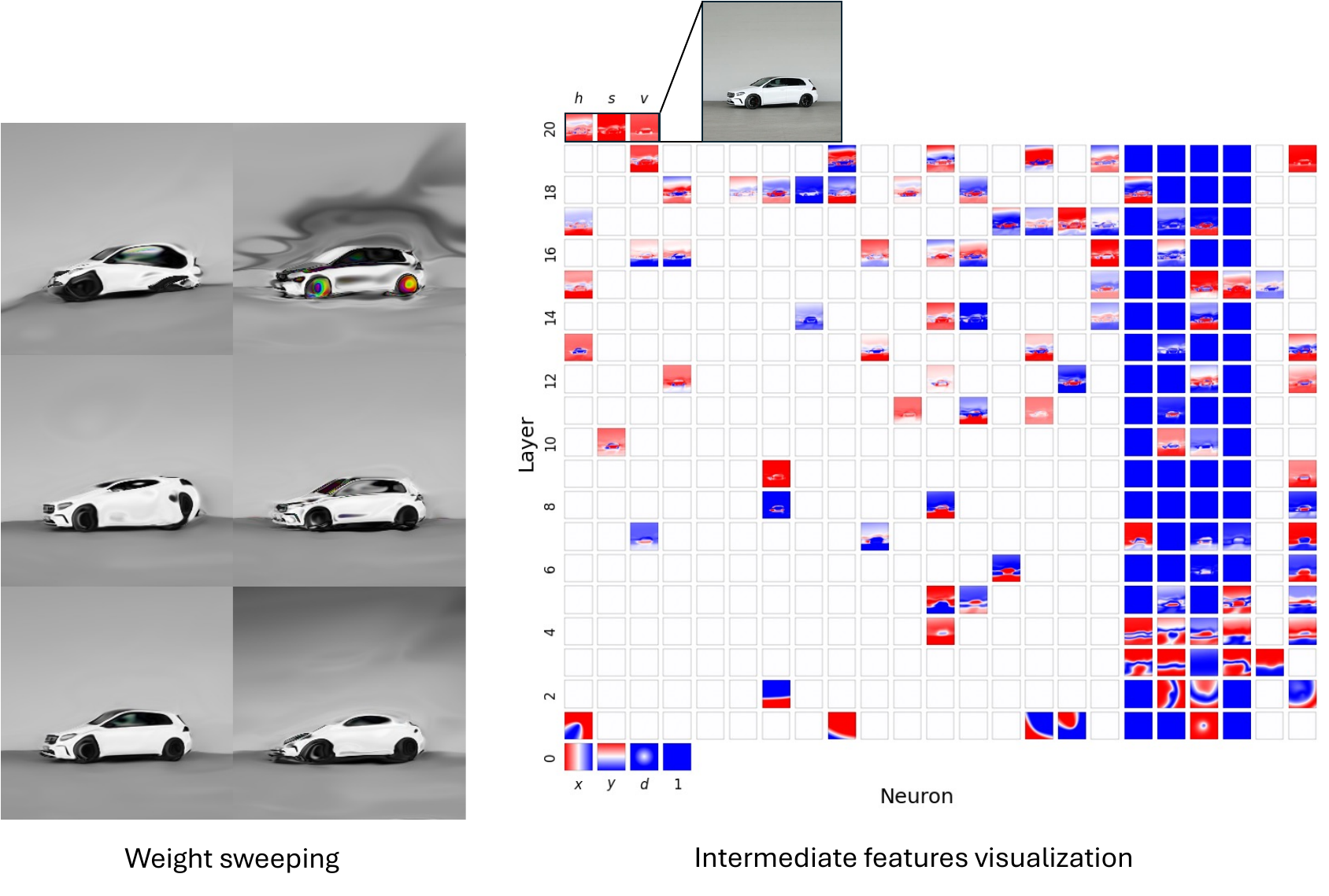}
    \caption{Full visualization of SP on a car image with some corresponding images from weights sweeping}
\end{figure*}

\begin{figure*}[!htb]
    \centering
    \includegraphics[width=\linewidth]{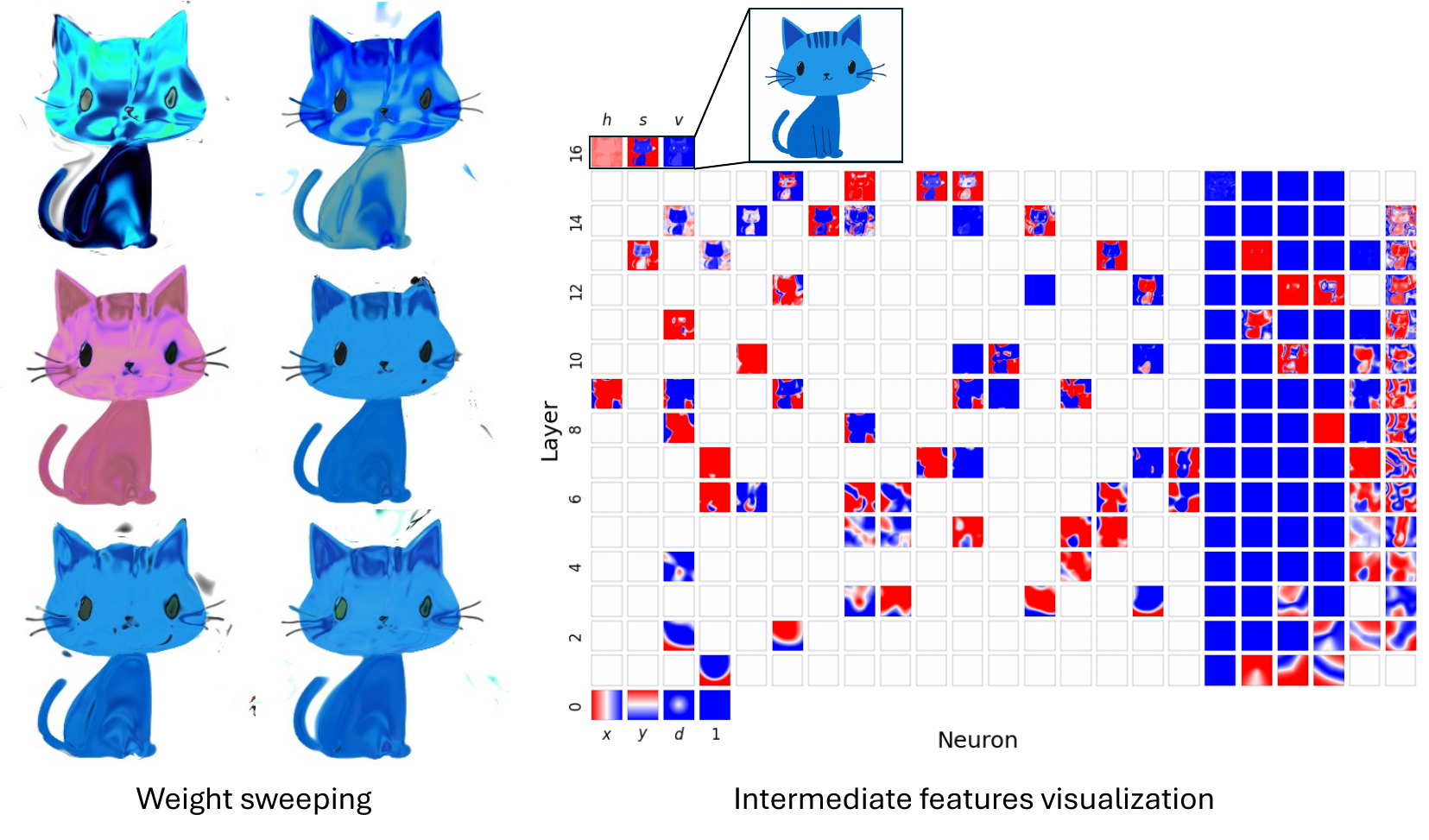}
    \caption{Full visualization of SP on a cat image with some corresponding images from weights sweeping.}
\end{figure*}

\begin{figure*}[!htb]
    \centering
    \includegraphics[width=0.75\linewidth]{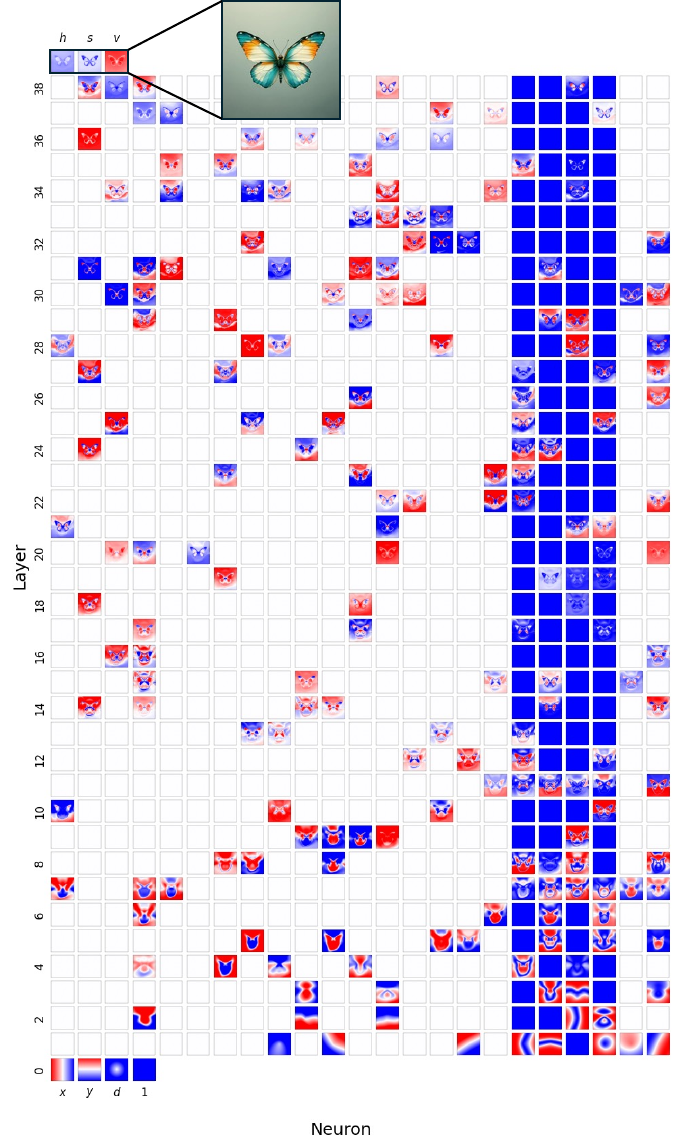}
    \caption{Full visualization of SP on a butterfly image.}
\end{figure*}

\begin{figure*}[!htb]
    \centering
    \includegraphics[width=\linewidth]{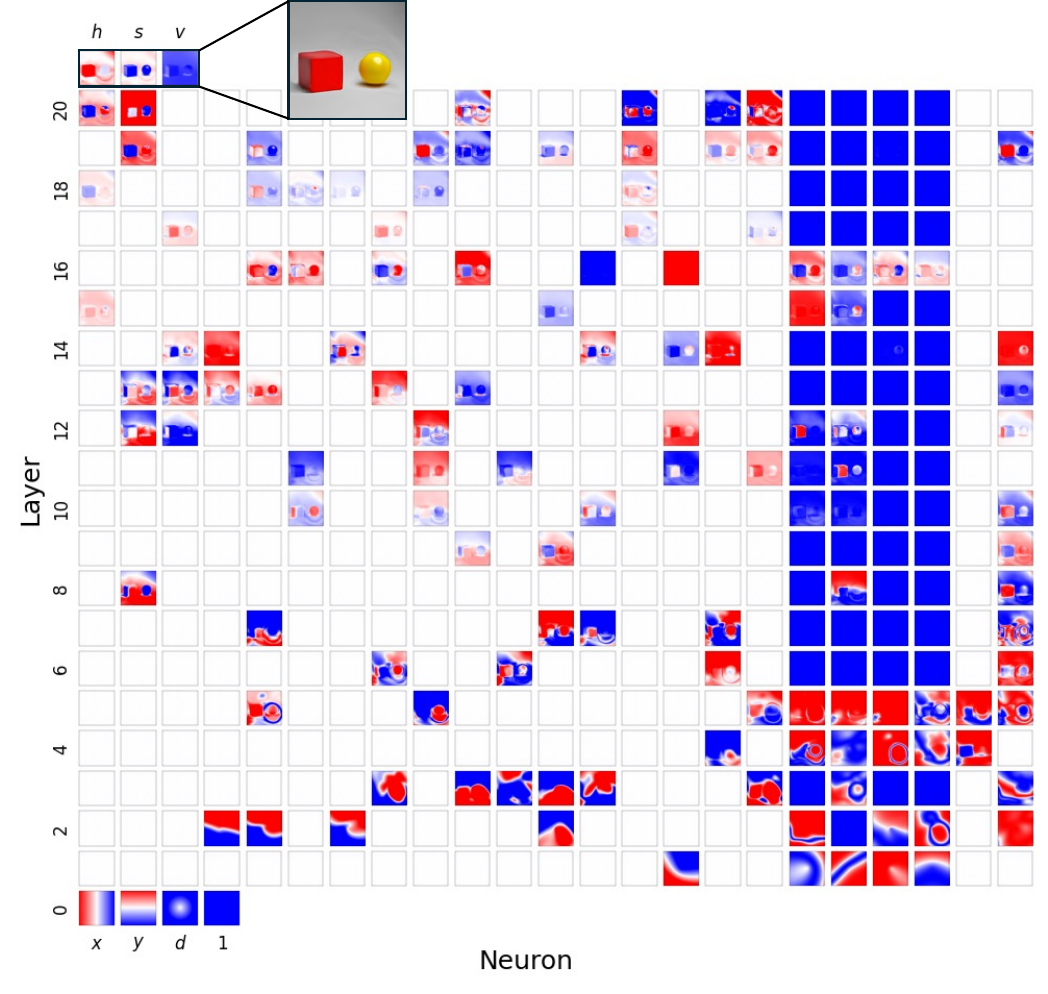}
    \caption{Full visualization of SP on an image illustrating a red cube and a yellow sphere.}
    \label{fig:spatial_compositionality}
\end{figure*}

\begin{figure*}[!htb]
    \centering
    \includegraphics[width=0.6\linewidth]{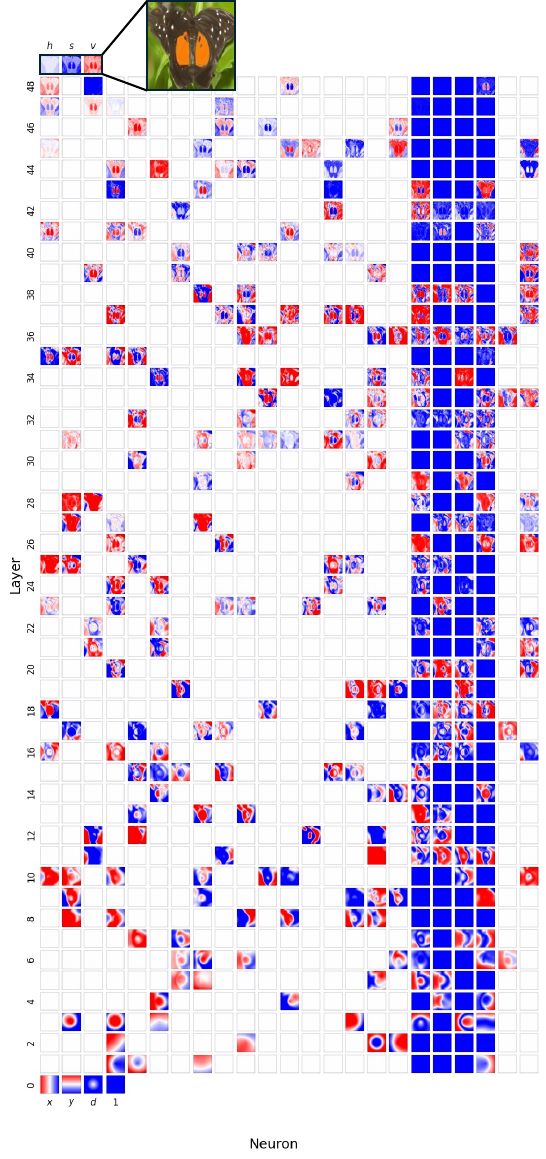}
    \caption{Full visualization of SP on an image illustrating a real butterfly with background.}
    \label{fig:background}
\end{figure*}

\begin{figure*}[!htb]
    \centering
    \includegraphics[width=0.6\linewidth]{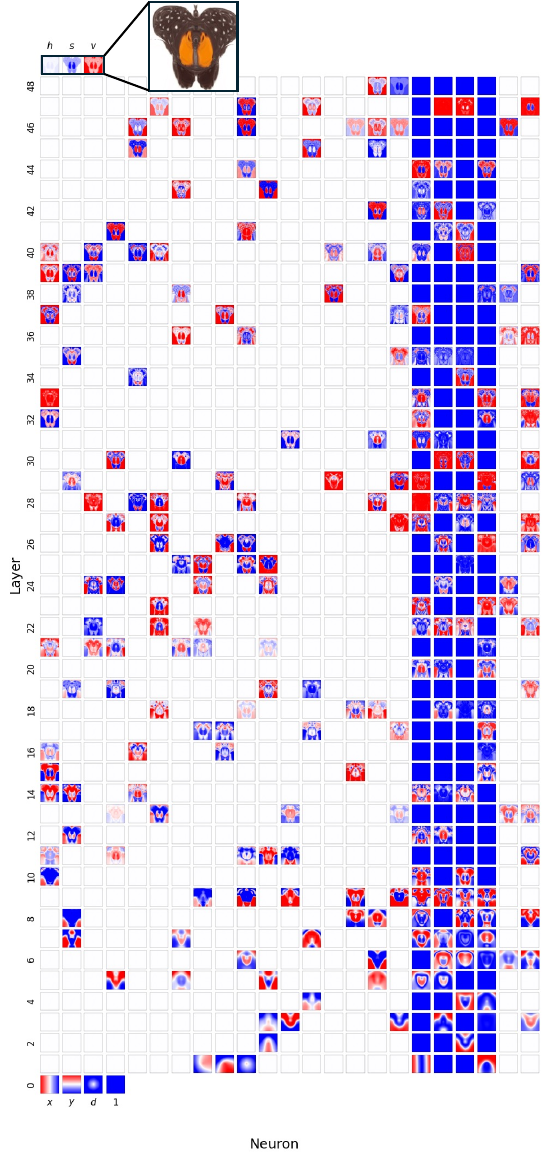}
    \caption{Full visualization of SP on an image illustrating a real butterfly without background.}
    \label{fig:nobackground}
\end{figure*}

\clearpage
\section{Other Pruning Results}
\label{ap:other_pruning}
In this section, we provide the visualization of internal representations, produced by using SP (round 1), and further visualize those produced by other pruning methods, reaching the same amount of parameters. In all the figures, SP yield clear symmetries and compositionality while Lottery Ticket Hypothesis \cite{frankle2018lottery}, LLM-Pruner \cite{ma2023llm}, Wanda \cite{sun2023simple}
\label{ap:prune}
\begin{figure*}[!htb]
    \centering
    \includegraphics[width=0.9\linewidth]{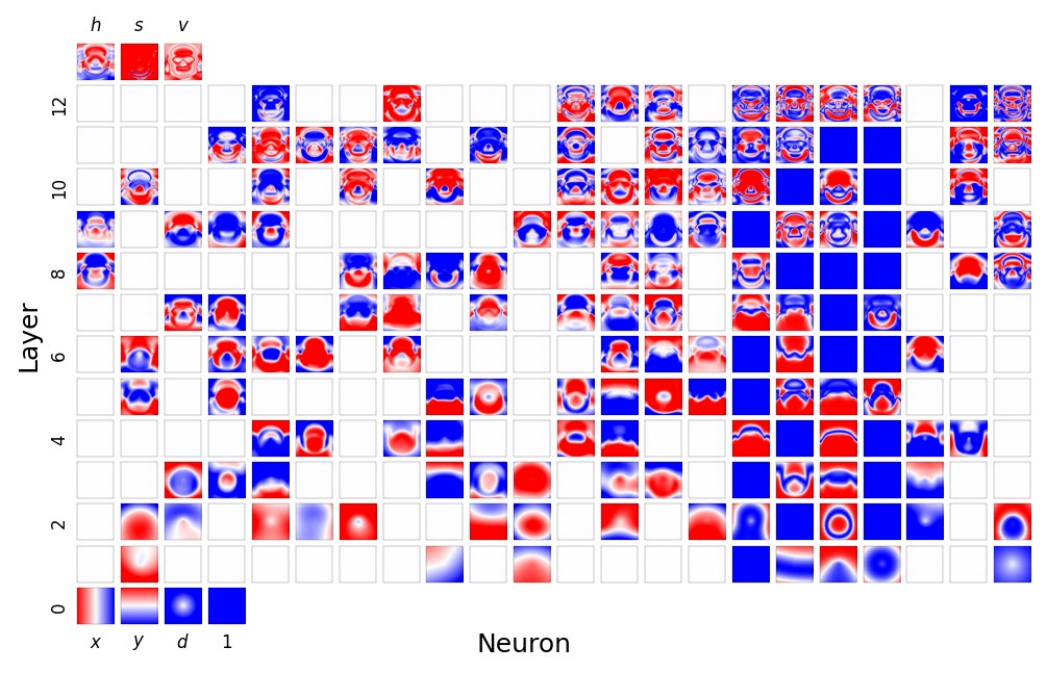}
    \caption{SP round 1 - 1404 parameters}
\end{figure*}

\begin{figure*}[!htb]
    \centering
    \includegraphics[width=0.9\linewidth]{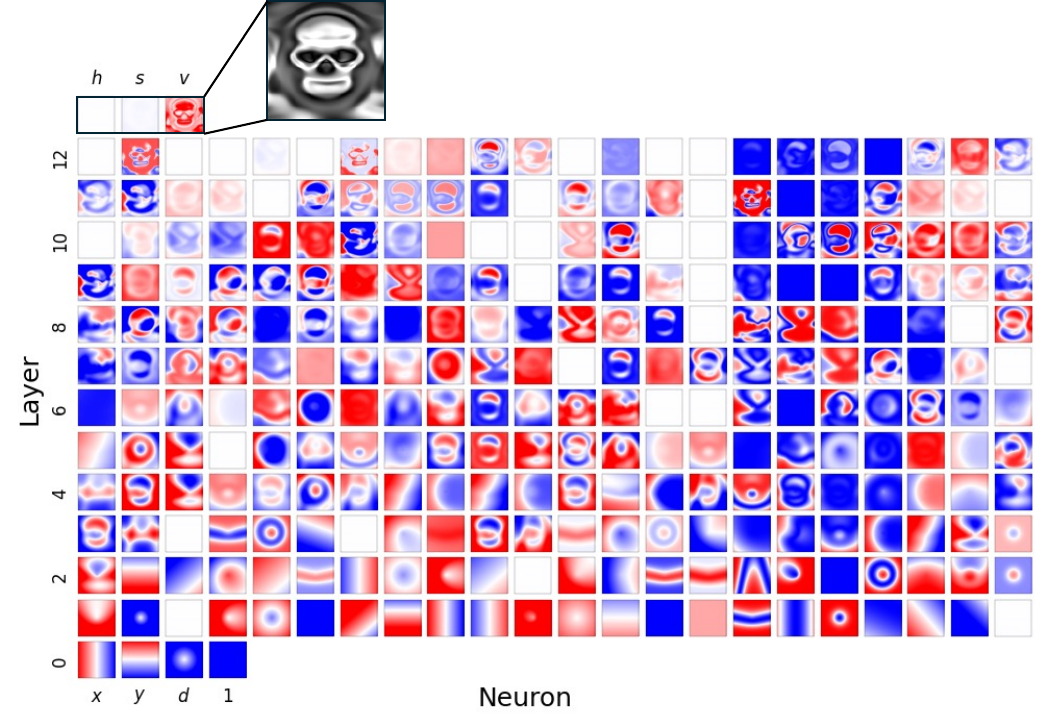}
    \caption{Lottery Ticket Hypothesis \cite{frankle2018lottery} on skull image, 473 weights}
\end{figure*}

\begin{figure*}[!htb]
    \centering
    \includegraphics[width=0.9\linewidth]{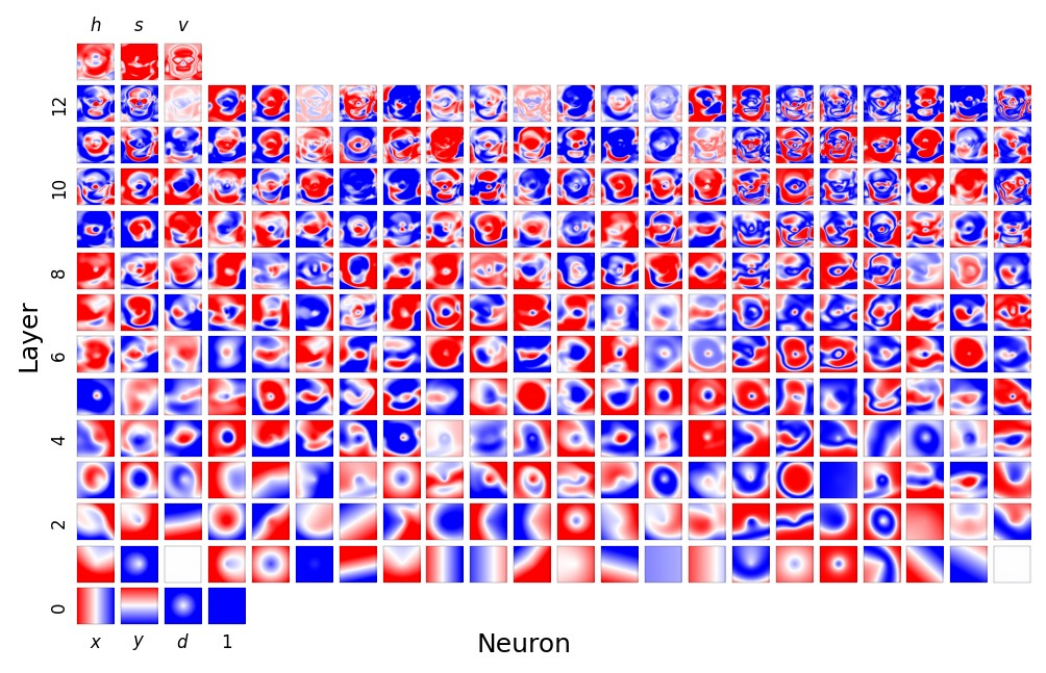}
    \caption{Lottery Ticket Hypothesis \cite{frankle2018lottery} on skulll image, 1452 weights}
\end{figure*}

\begin{figure*}[!htb]
    \centering
    \includegraphics[width=0.9\linewidth]{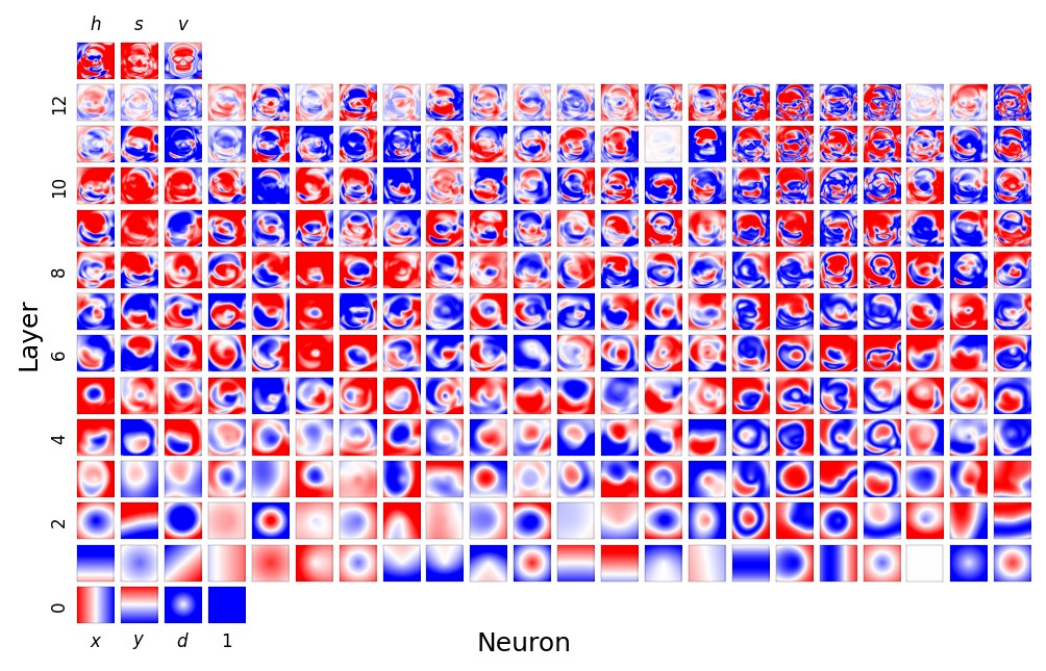}
    \caption{Wanda \cite{sun2023simple} on skull image, 1404 weights}
\end{figure*}

\begin{figure*}[!htb]
    \centering
    \includegraphics[width=0.9\linewidth]{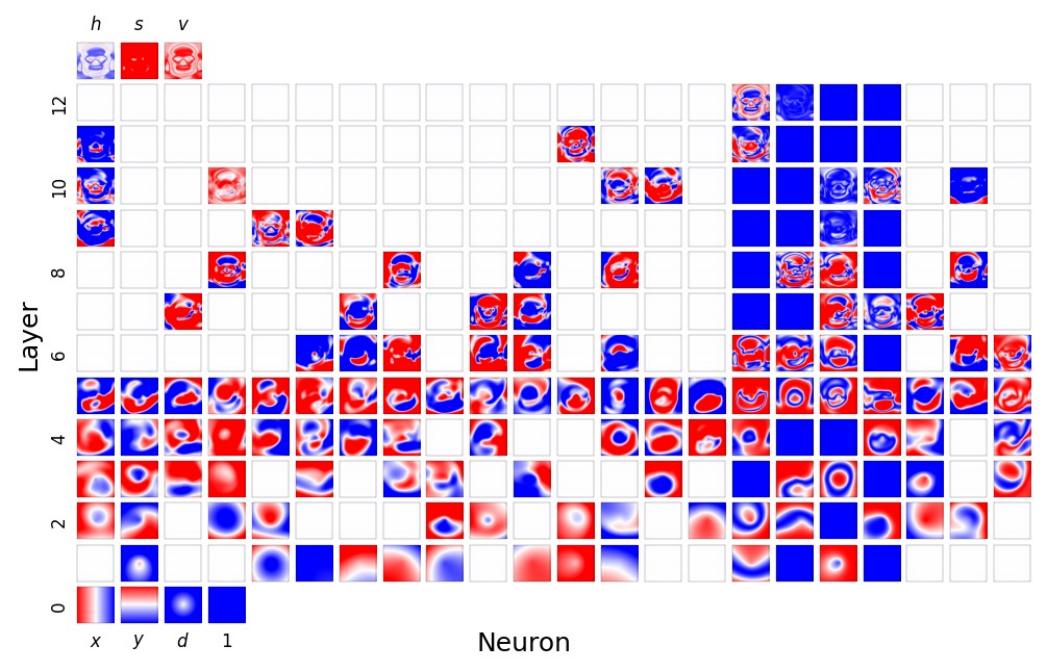}
    \caption{LLM-Pruner \cite{ma2023llm} on skull image, 1397 weights}
\end{figure*}

\begin{figure*}[!htb]
    \centering
    \includegraphics[width=\linewidth]{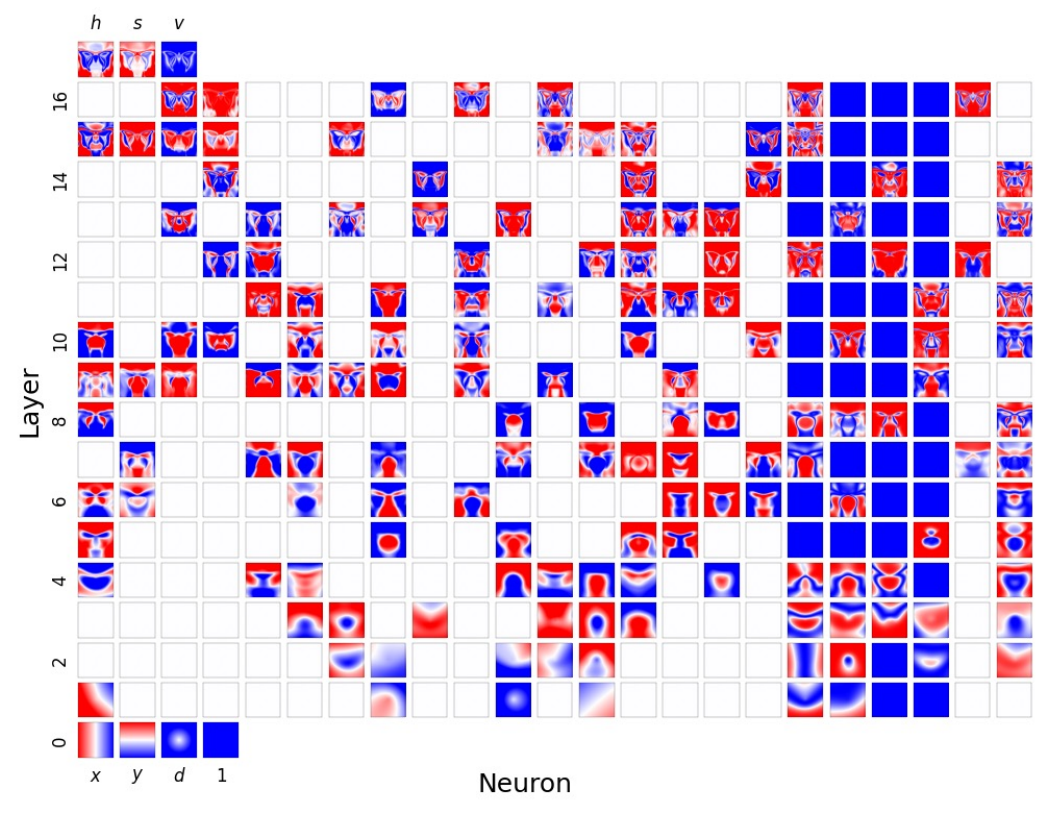}
    \caption{SP round 1 on butterfly image, 3405 weights}
\end{figure*}

\begin{figure*}[!htb]
    \centering
    \includegraphics[width=\linewidth]{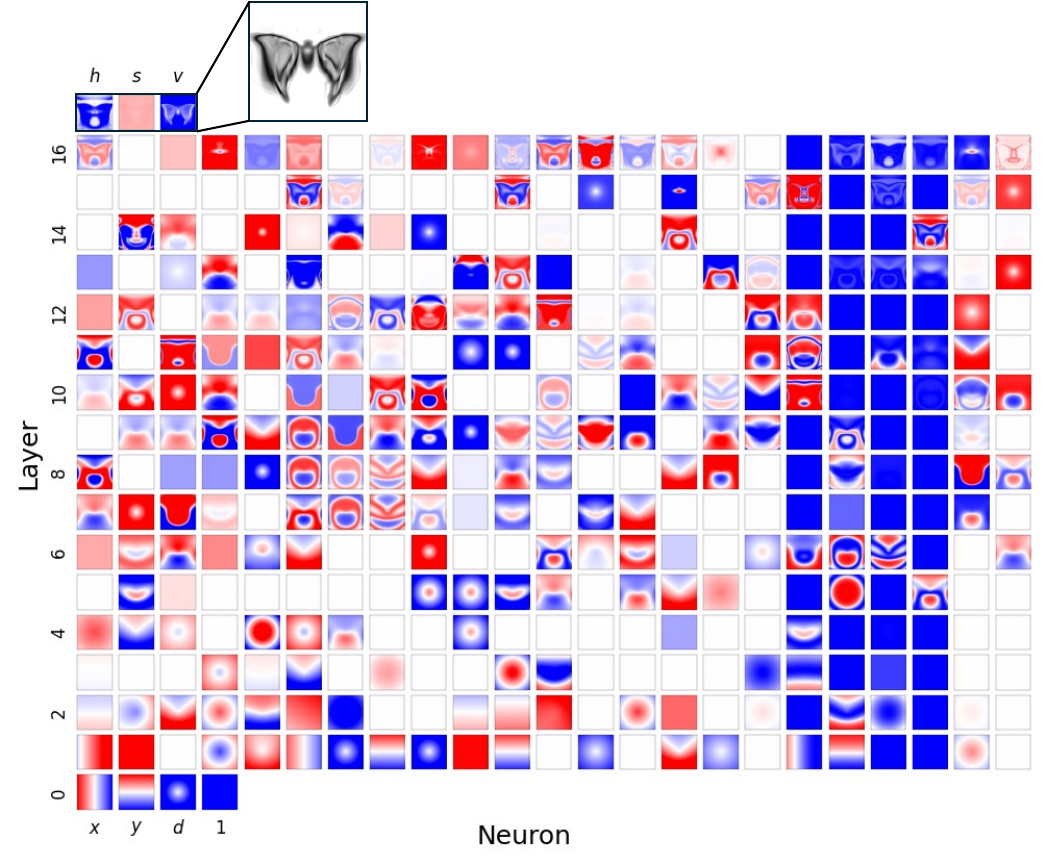}
    \caption{Lottery Ticket Hypothesis \cite{frankle2018lottery} on butterfly image, 468 weights}
\end{figure*}

\begin{figure*}[!htb]
    \centering
    \includegraphics[width=\linewidth]{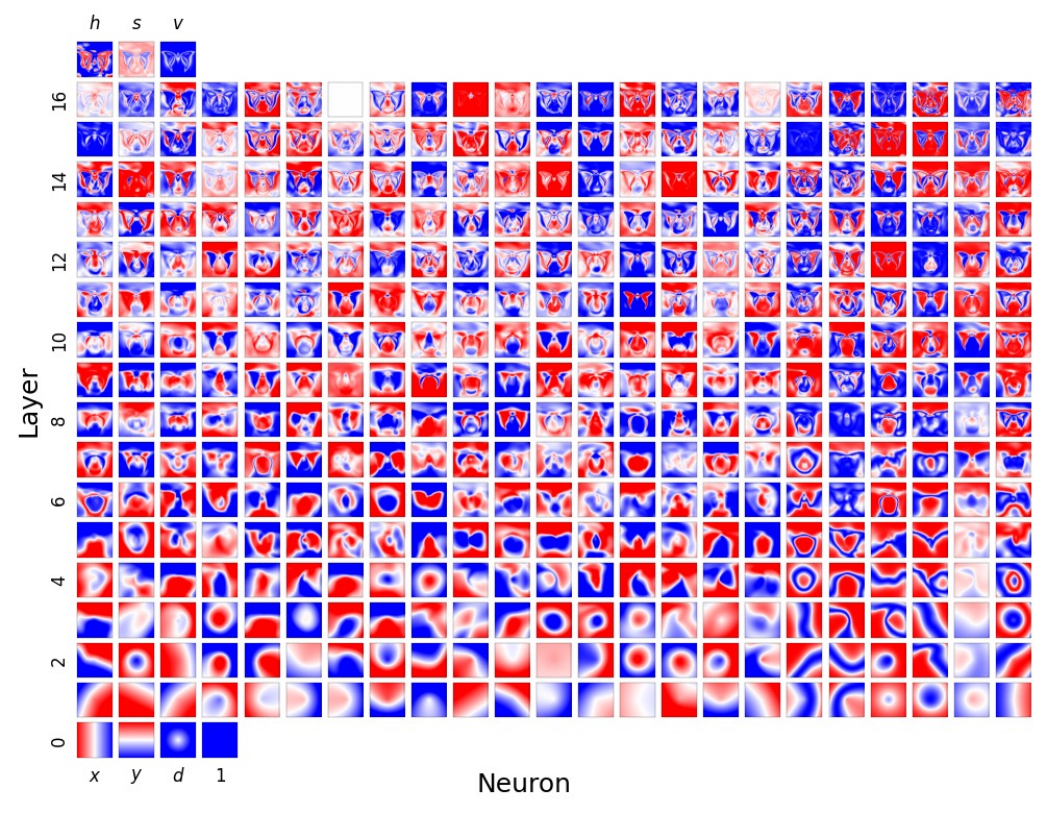}
    \caption{Lottery Ticket Hypothesis \cite{frankle2018lottery} butterfly image, 3392 weights}
\end{figure*}

\clearpage
\section{Ablation study on learning rate}

We plot the training curves across three optimizers' settings and 4 learning rates.
\label{ap:lr}
\begin{figure*}[!htb]
    \centering
    \includegraphics[width=\linewidth]{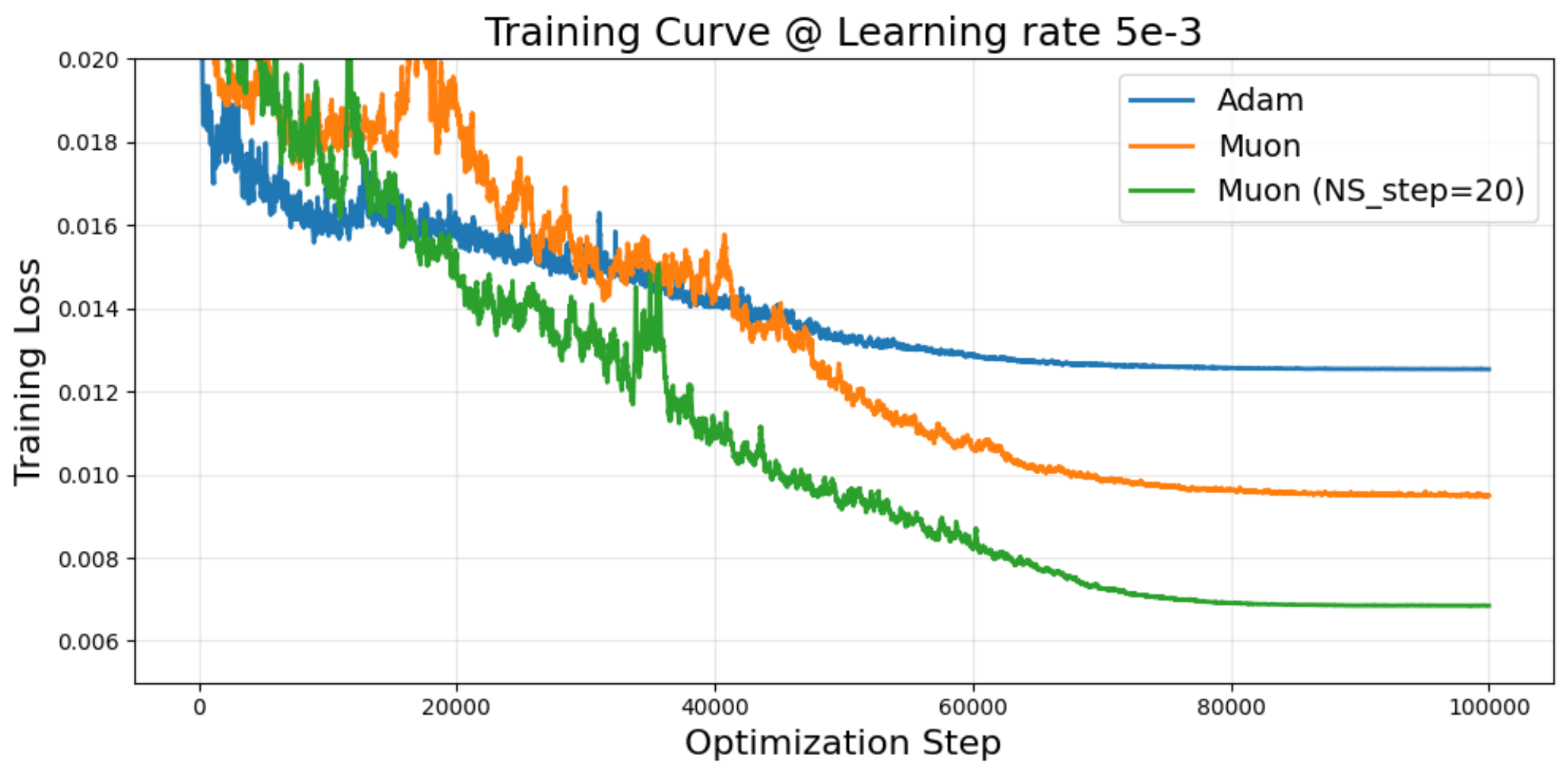}
    \caption{Training loss with different optimizers using learning rate 5e-3}
\end{figure*}

\begin{figure*}[!htb]
    \centering
    \includegraphics[width=\linewidth]{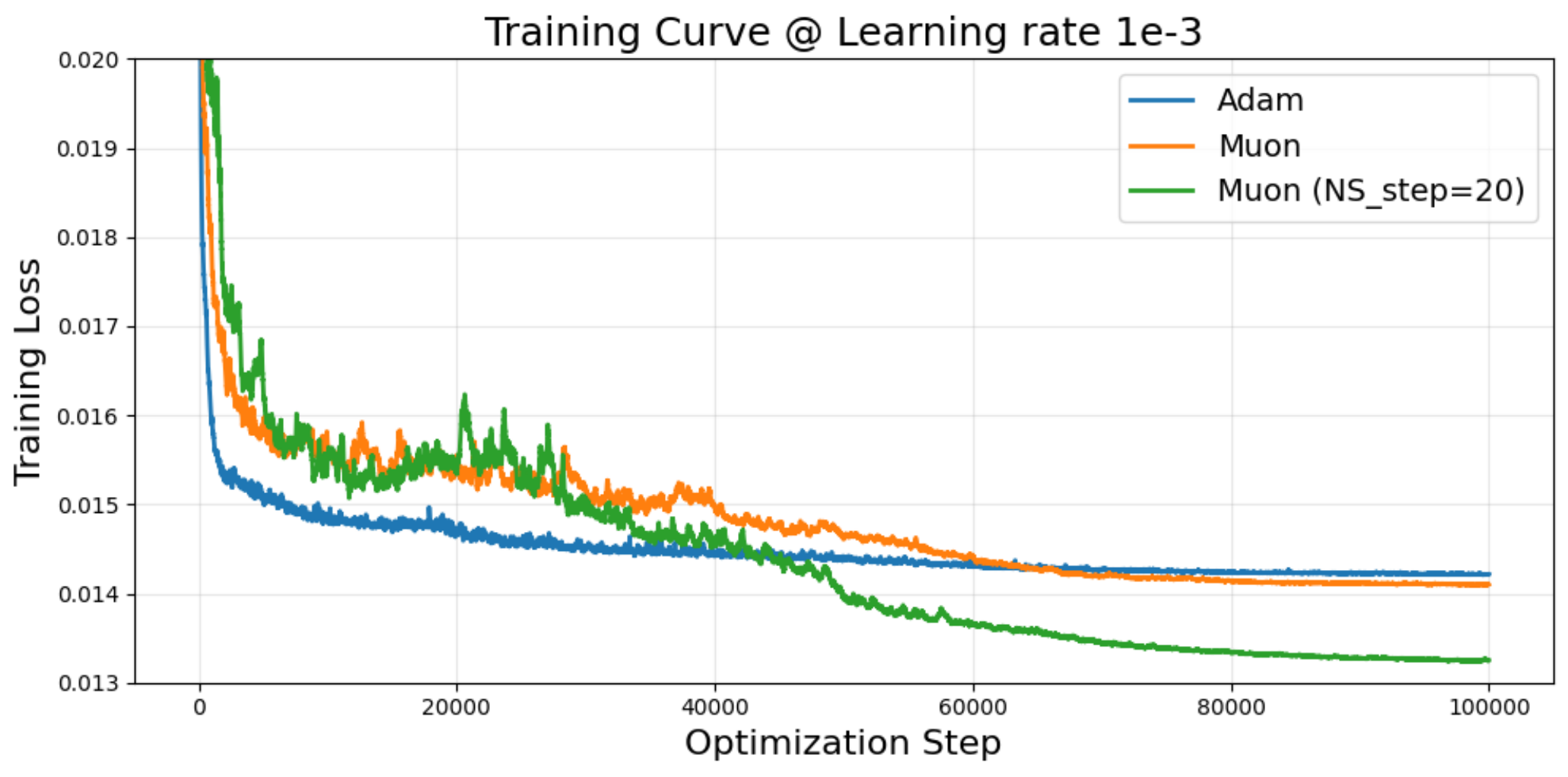}
    \caption{Training loss with different optimizers using learning rate 1e-3}
\end{figure*}

\begin{figure*}[!htb]
    \centering
    \includegraphics[width=\linewidth]{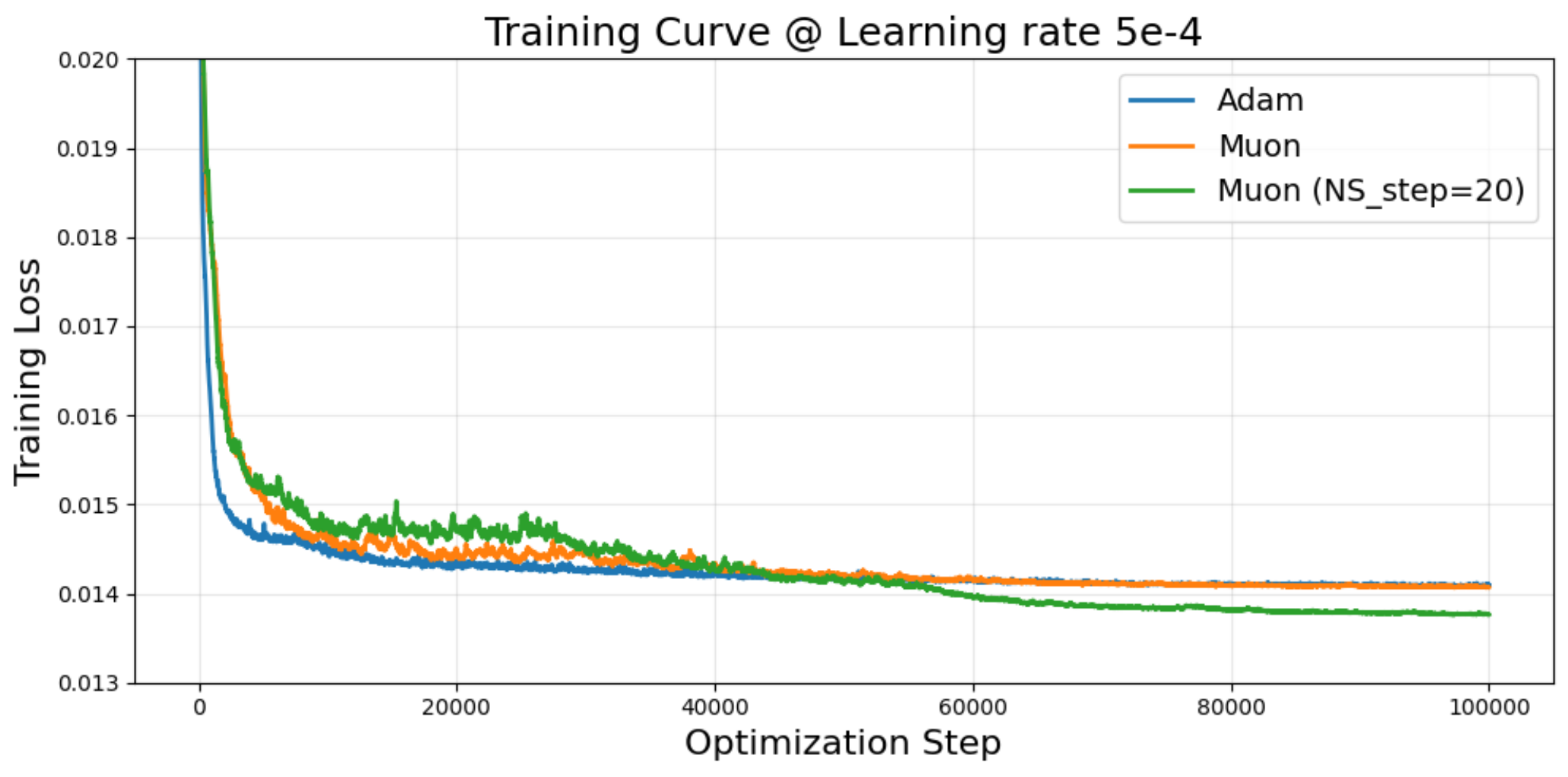}
    \caption{Training loss with different optimizers using learning rate 5e-4}
\end{figure*}

\begin{figure*}[!htb]
    \centering
    \includegraphics[width=\linewidth]{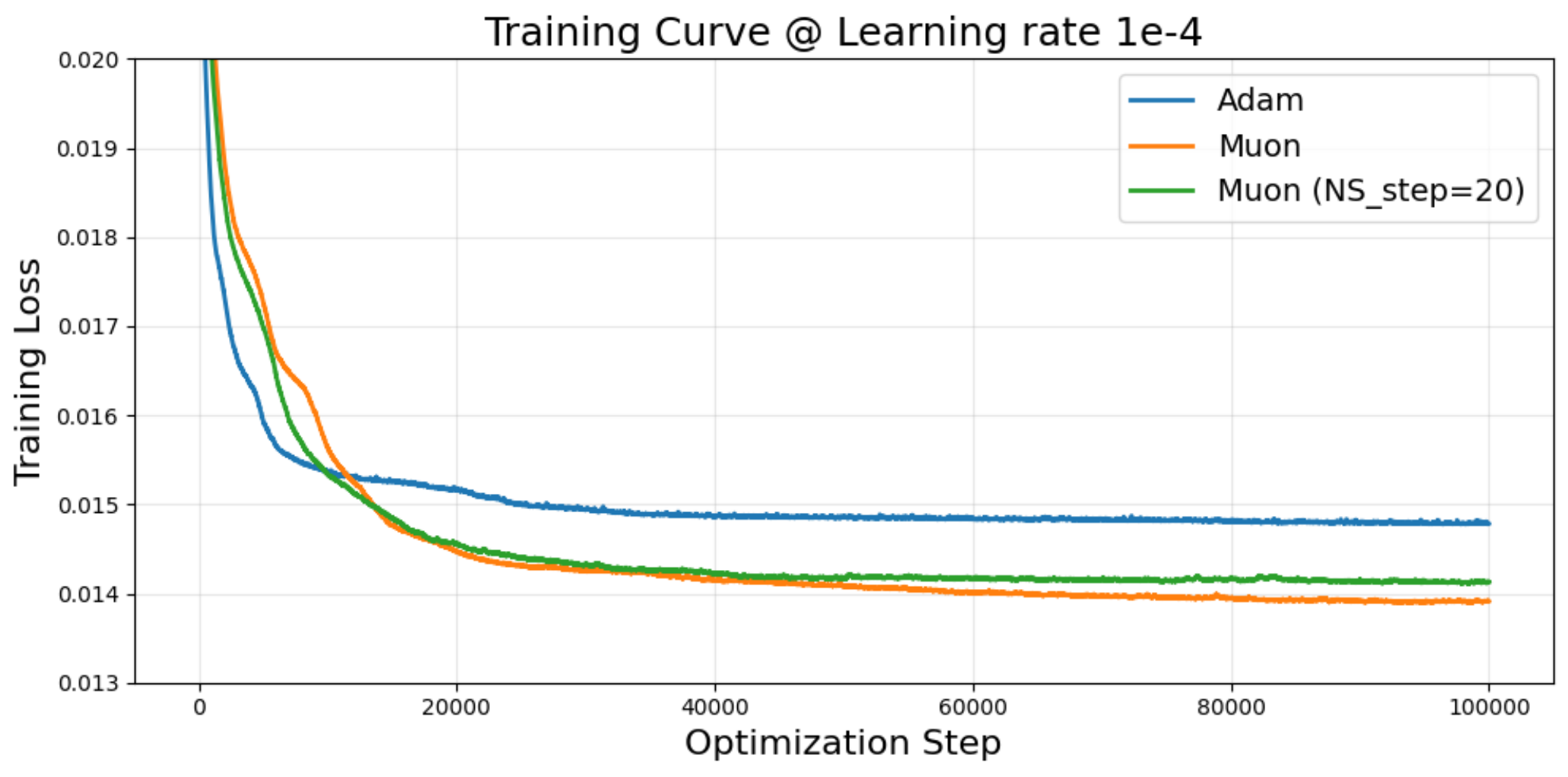}
    \caption{Training loss with different optimizers using learning rate 1e-4}
\end{figure*}

\clearpage

\section{Multi-images Training}
\label{ap:multi}
\begin{figure*}[!h]
    \centering
    \includegraphics[width=\linewidth]{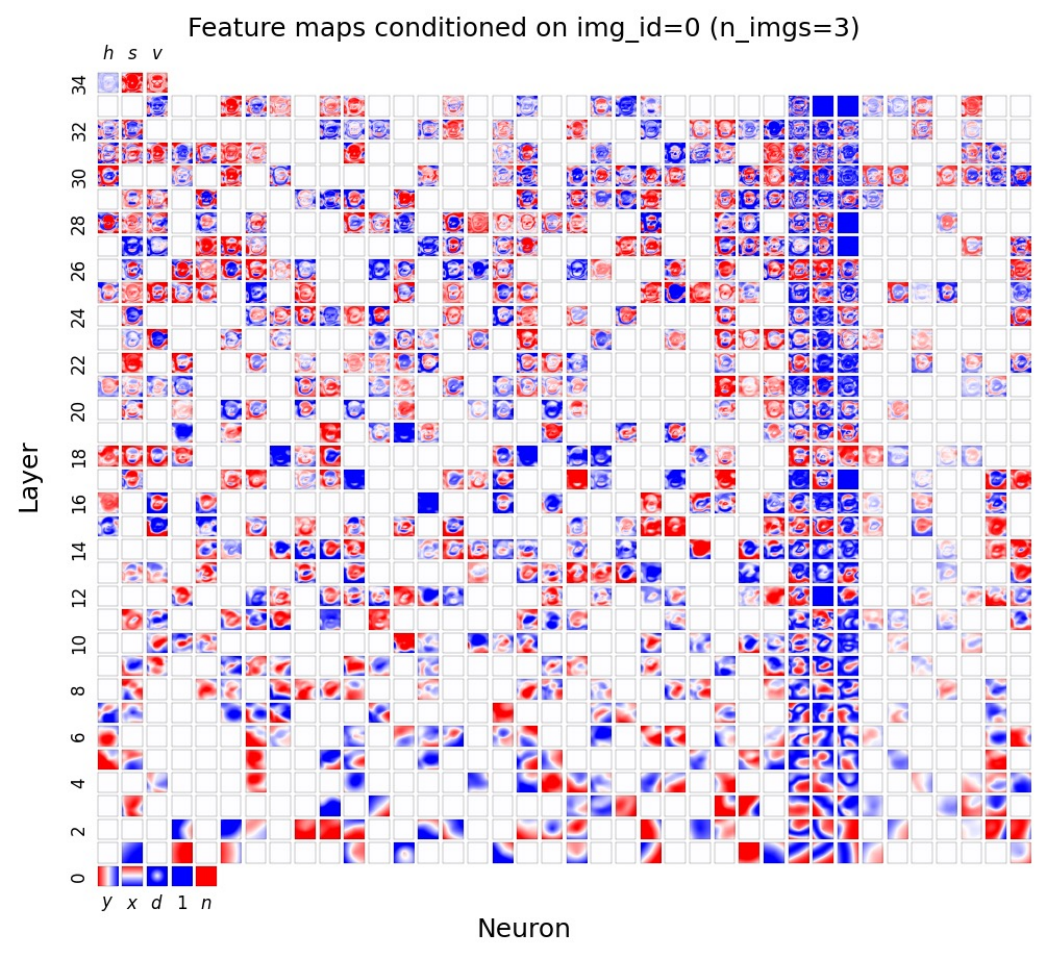}
    \caption{Multi-CPPN for Picbreeder's skull, $n$=-1}
    \label{multi_skull}
\end{figure*}

\begin{figure*}
    \centering
    \includegraphics[width=\linewidth]{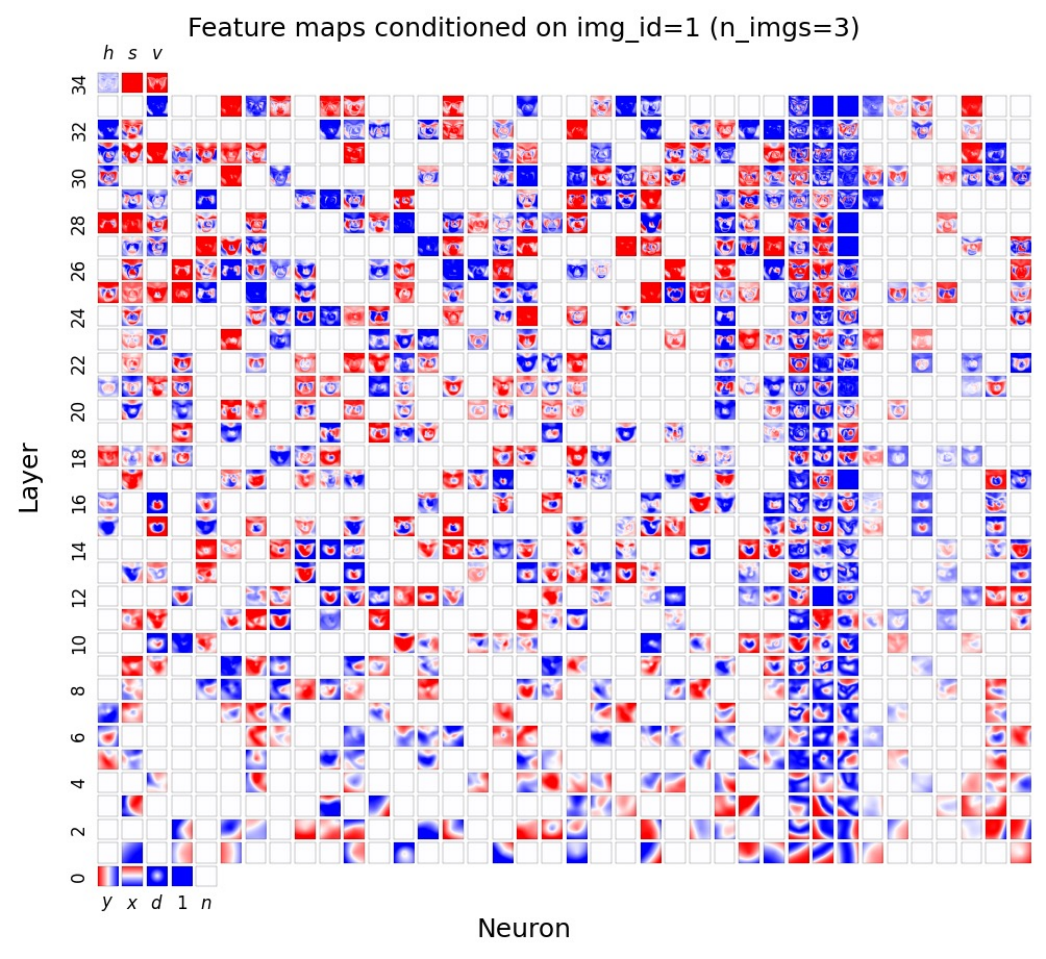}
    \caption{Multi-CPPN for Picbreeder's butterfly, $n$=0}
    \label{multi_butterfly}
\end{figure*}

\begin{figure*}
    \centering
    \includegraphics[width=\linewidth]{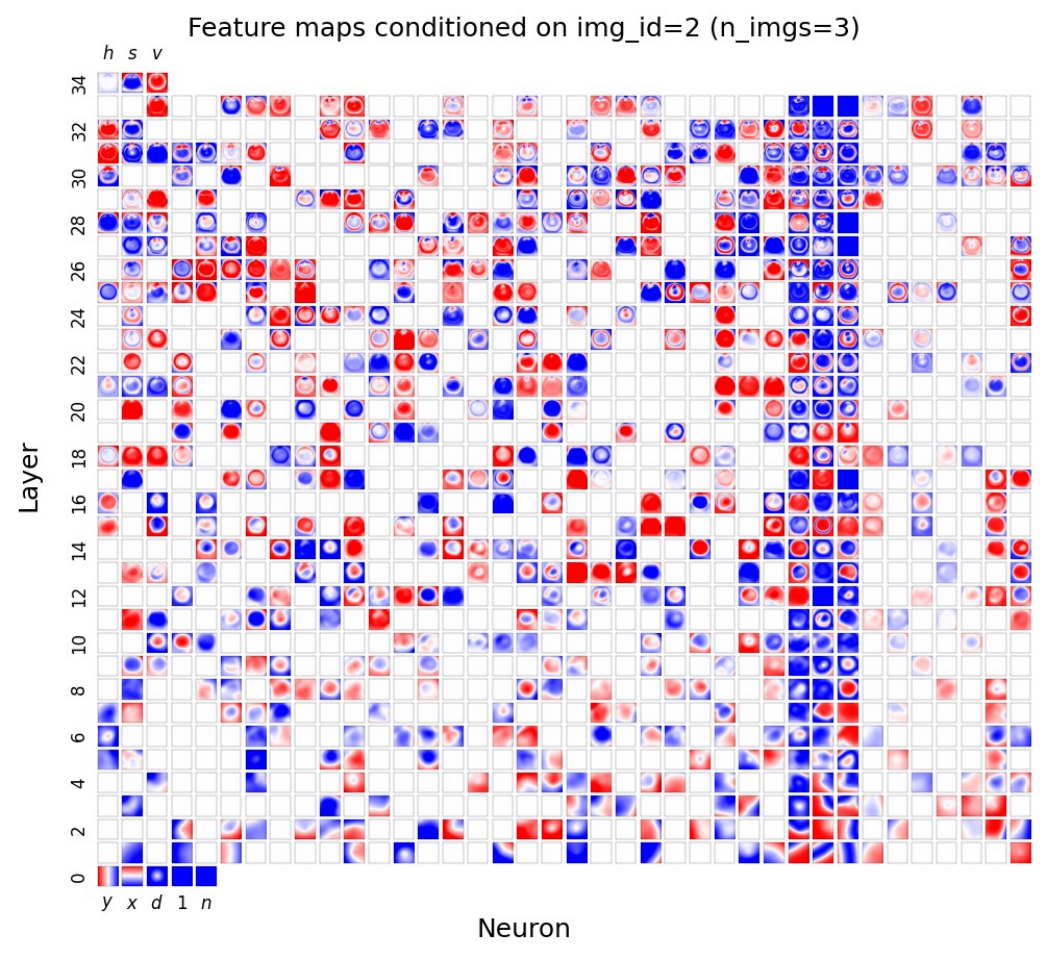}
    \caption{Multi-CPPN for Picbreeder's apple, $n$=1}
    \label{multi_apple}
\end{figure*}

In this section, we try to adopt the CPPN $F_\theta$ to be a multi-CPPN: $(h_n,s_n,v_n)=F_\theta(x,y,d,1,n)$ with $n$ is the normalized index of the image to fit. To train on 3 Picbreeder's artifacts: skull, butterfly, and apple, the corresponding $n=-1, 0, 1$, respectively. We apply the SP but observe no compositionality. Figures~\ref{multi_skull},~\ref{multi_butterfly},~\ref{multi_apple} visualize the internal features.

